\documentclass[a4paper]{article}
\usepackage[utf8]{inputenc}
\pdfoutput=1

\usepackage{amsmath}

\title{Forest Floor Visualizations of Random Forests}

\usepackage{lineno,hyperref}
\modulolinenumbers[5]

\usepackage{breqn} 
\usepackage[english]{babel}
\usepackage{grffile} 
\usepackage{amssymb} 
\usepackage{graphicx}

\usepackage{graphicx}

\usepackage{caption}
\usepackage{subcaption}
\usepackage{geometry}
\usepackage{multicol}

\usepackage{authblk}

\author[1,2]{Soeren H. Welling}
\author[2]{Hanne H.F. Refsgaard}
\author[1]{Per B. Brockhoff}
\author[1]{Line H. Clemmensen\thanks{lkhc@dtu.dk}}
\affil[1]{Department of Applied Mathmatics and Computer Science, Technical University of Denmark, Matematiktorvet, Building 324, 2800 Kgs. Lyngby, Denmark}
\affil[2]{Novo Nordisk Global Research, Novo Nordisk Park 1, 2760 Maaloev, Denmark}

\begin{document}
\maketitle
\begin{abstract}

We propose a novel methodology, forest floor, to visualize and interpret random forest (RF) models. RF is a popular and useful tool for non-linear multi-variate classification and regression, which yields a good trade-off between robustness (low variance) and adaptiveness (low bias). Direct interpretation of a RF model is difficult, as the explicit ensemble model of hundreds of deep trees is complex. Nonetheless, it is possible to visualize a RF model fit by its mapping from feature space to prediction space. Hereby the user is first presented with the overall geometrical shape of the model structure, and when needed one can zoom in on local details. Dimensional reduction by projection is used to visualize high dimensional shapes. The traditional method to visualize RF model structure, partial dependence plots, achieve this by averaging multiple parallel projections. We suggest to first use feature contributions, a method to decompose trees by splitting features, and then subsequently perform projections. The advantages of forest floor over partial dependence plots is that interactions are not masked by averaging. As a consequence, it is possible to locate interactions, which are not visualized in a given projection. Furthermore, we introduce: a goodness-of-visualization measure, use of colour gradients to identify interactions and an out-of-bag cross validated variant of feature contributions.

\end{abstract}

\begin{multicols}{2}

\section{Introduction}

We propose a new methodology, forest floor, to visualize regression and classification problems through feature contributions of decision tree ensembles such as random forest (RF). Hereby, it is possible to visualize an underlying system of interest even when the system is of higher dimensions, non-linear, and noisy. 2D or 3D visualizations of a higher-dimensional structure may lead to details, especially interactions, not being identifiable. Interactions in the model structure mean that the model predictions in part rely on the interplay on two or more features. Thus, the interaction parts of a model structure cannot be reduced to additive scoring rules, one for each feature. Likewise, to plot single feature-to-prediction relationships is not a sufficient context for visualizing any interactions. Often a series of complimentary visualizations are needed to produce an adequate representation. It can be quite time consuming to look through any possible low dimensional projection of the model structure to check for interactions. Forest floor guides the user in order to locate prominent interactions in the RF model structure and to estimate how influential these are.

For RF modeling, hyper parameter tuning is not critical and default parameters will yield acceptable model fits and visualizations in most situations \cite{Liaw2002,Svetnik2003}. Therefore, it is relatively effortless to train a RF model. In general, for any system where a model has a superior prediction performance, it should be of great interest to learn its model structure. Even within statistical fields, where decision tree ensembles are far from standard practice, such insight from a data driven analysis can inspire how to improve goodness-of-fit of a given model driven analysis.

Although the RF algorithm by Breimann \cite{Breimann2001} has achieved the most journal citations, other later decision tree ensemble models/algorithms such as ExtraTrees \cite{Maree2005}, conditional inference forest \cite{Hothorn2006}, Aborist \cite{Seligman2015}, Ranger \cite{Ranger2015} and sklearn.random.forest \cite{sklearn} will often outperform the original RF on either prediction performance and/or speed. These models/algorithms differ only in their software implementation, split criterion, agreggation or in how deep the trees are grown. Therefore all variations are compatible with the forest floor methodology. Another interesting variant, rotation forest \cite{Rodriguez2006}, does not make univariate splits and is therefore unfortunately not directly compatible with forest floor visualizations. To expand the use of feature contributions and forest floor, we also experimented with computing feature contributions for gradient boosted trees \cite{friedman2001}. This is possible, as splits still are univariate and trees contribute additively to the ensemble prediction. A proof-of-concept of computing feature contributions on gradient boosted regression trees and visualizations are provided in supplementary materials.

Decision trees, as well as other machine learning algorithms, such as support vector machines and artificial neural networks can fit regression and classification problems of complex and noisy data, often with a high prediction performance evaluated by prediction of test sets, n-fold cross validation, or out-of-bag (OOB) cross validation. The algorithms yield data driven models, where only little prior belief and understanding is required. Instead, a high number of observation are needed to calibrate the adaptive models. The models themselves are complex black-boxes and can be difficult to interpret. If a data driven model can reflect the system with an impressive prediction performance, the visualization of the model may deduce knowledge on how to interpret the system of interest. In particular, a good trade-off between generalization power and low bias is of great help, as this trade-off in essence sets the boundary for what is signal and what is noise. The found signal is the model fit, which can be represented as the mapping from feature space to prediction space (output, target, response variable, dependent variable, y). The noise is the residual variance of the model. The estimated noise component will both be due to random/external effects but also lack of fit.\par

\subsection{Overview of the article}
In this article we introduce the forest floor methodology. The central part is to define a new mapping space visualization, forest floor. Forest floor rely on the feature contributions method \cite{Kuz'min2011}\cite{Palczewska2014}, rather than averaging many projections (partial dependence) \cite{friedman2001} or projecting the average (sensitivity analysis) \cite{Cortez2013}. In Section \ref{intrRepRF} these previous mapping space visualizations are introduced and the challenges to overcome are discussed.
In the theory section, \ref{TheoDefMap}, we discuss the feature space, prediction space and the joined mapping space for any regression or classification model and define local increments as vectors in the prediction space. Properties of the RF algorithm by Breimann  \cite{Breimann2001} and the feature contributions method by Kuz'min \textit{et al} \cite{Kuz'min2011} and Palczewska \textit{et al} \cite{Palczewska2014} are highlighted and illustrated in section \ref{TheoRF}. In section \ref{TheoLI} we argue that the prediction of any node in any tree is a point in the prediction space and the local increments are the vectors that connect the nodes of the trees. Any prediction for any observation is basically a summed sequence of local increments plus the grand mean or base rate. Since local increments are vectors and not a tree graph, the sum of vectors is not dependent on the order of the sequence. In Section \ref{TheoDecompose} we show how that feature contributions, a particular reordering of local increments by splitting feature, can be used to decompose the model structure \ref{TheoDecompose}.
We also introduce a new cross-validated variant of feature contributions and provide an elaborated definition of feature contribution to also account exactly for the bootstrapping process and/or stratification.

The materials and methods sections, \ref{materialDataSoft} and \ref{materialToyData}, provide instructions on how to reproduce all visualization in this paper. The result section \ref{resStart} is dedicated to three practical examples of visualizing models with forest floor. The three examples are a simulated toy data set, a regression problem (white whine quality) and a classification problem (contraception method choice). A low-dimensional visualization is not likely to convey all aspects of a given RF mapping surface. For all practical examples, we describe how to find an adequate series of visualizations that do.

\subsection{Representations of random forest models}
\label{intrRepRF}
A RF model fit, like other decision tree based models, can be represented by the graphs of the multiple trees. Few small tree graphs can be visualized and comprehended. However, multiple fully grown trees are typically needed to obtain an optimal prediction performance. Such a representation cannot easily be comprehended and is thus inappropriate for interpretation of model fits. A random forest fit can be seen as a large set of split rules which can be reduced to a smaller set of simpler rules, when accepting a given increase in bias. This approach has been used to reduce the model complexity \cite{Liu2014}. But if the minimal set of rules still contains a large number, e.g. hundreds or thousands, then this simplified model fit is still incomprehensible. It is neither certain which rules have influence on predictions nor which rules tend to cancel each other out. We believe that the rule-set or tree-structure representations are mainly appropriate to understand how a RF algorithm possibly can model data. On the other hand, these representations are indeed inappropriate for interpreting RF model fits and conveying the overall model structure. For that purpose, a mapping space visualization is superior in terms of visualization and communication.

\begin{figure*}[htp]
\centering
\includegraphics{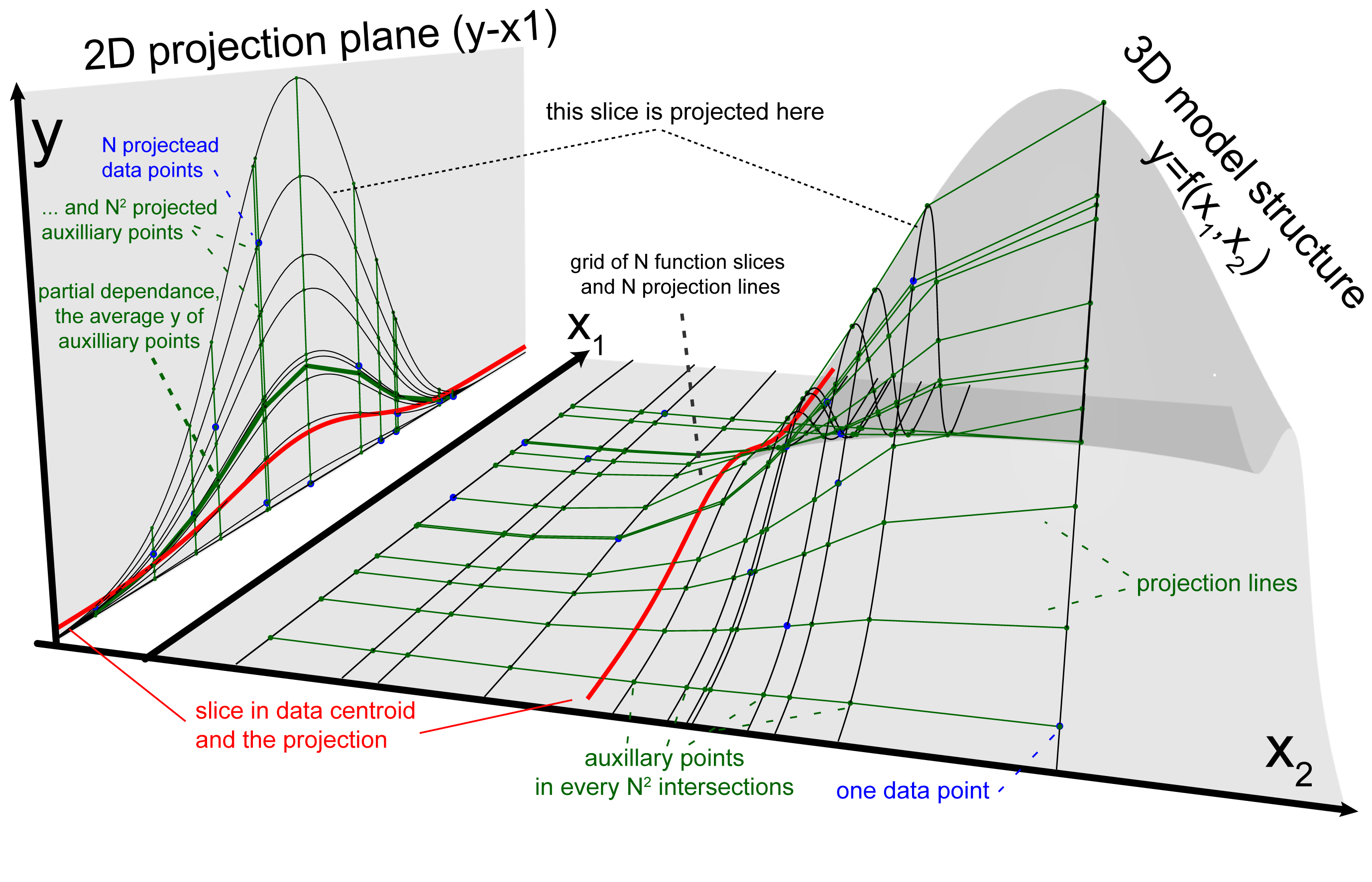}
\caption{Illustration of sensitivity analysis and partial dependence plots. The grey response surface depicts a given learned model structure of two input features ($X_1$ and $X_2$) and one prediction axis ($\hat{y}$). 11 data points vs. predictions are depicted as blue dots. 1D-sensitivity analysis (fat red lines): one partial function slice intersects the centroid where $X_2 = \overline{X}_2$ an is projected to the $X_1$-$y$ plane. ICE plot: Multiple function slices (black lines) all parallel to $X-1$ intersects each one data point and all slices are projected to the $X_1$-$y$ plane. Partial dependence plots: Each data point intersected by one black line is projected to any black lines (green points). The green point outline a grid. All green and blue points are projected into the $X_1$-$y$ plane, and the fat green line connects the average prediction values as a function of $X-1$. This illustration can be generalized to any dimensional reduction.
}

\label{partialFigure}
\end{figure*}

If we join the feature space and prediction space, this function will be represented as a geometrical shape of points. Each point represents one prediction for a given feature combination. This geometrical shape is the model structure and is an exact representation of the model itself. Nevertheless, for a given \emph{d}-dimensional problem where $d>3$, this is still difficult to visualize or even comprehend. Instead, one may project/slice or decompose the high-dimensional mapping into a number of marginal visualization where small subsets of features can be investigated in turns. This allows us to comprehend the isolated interplay of one or a few features in the model structure.

Following, we will introduce previous examples of mapping space visualizations to specify what forest floor aims to improve. Different types of sensitivity analysis (SA) were used by Cortez and Embrechts to make such investigations \cite{Cortez2013}, we will here discuss sensitivity analysis and data based sensitivity analysis. First a supervised machine learning model is trained. Next the model is probed. That means to input a set of simulated feature observations (points in feature space) into the model fit and record the output (target predictions). Instead of probing the entire high-dimensional mapping space, only one confined slice of fewer dimensions is probed in order to make feasible visualizations.

The simplest visualization in SA is one dimensional (1D-SA), where a single feature is varied in a range of combinations, and this range will span the X-axis of the visualization. When two features are varied (2D-SA), the resulting grid of combinations will span the XY-plane. All other features must be fixed at e.g. the mean value, the feature centroid of the training set. The model fit is probed with these observations and the resulting predictions will be plotted by the Z-axis. The obtained line/surface will now visualize one particular 2D or 3D slice of the full mapping structure.

In figure \ref{partialFigure}, a non-linear regression model structure ($y = sin(X_1)^8 sin(X_2)^8 + \epsilon$) is represented by the grey transparent surface. The model has two feature axes in the horizontal XY-plan and the prediction axis by the vertical Z-axis. Thus, the mapping space has 3 dimensions and the model structure is some curved 2D-surface which connect any given feature combination with one prediction. The red line/slice in the model structure is the example of an 1D-SA visualization. This single slice is projected into the $X_1$-$Z$ plane. This 1D-SA projection portrays the partial effect of feature $X_1$ in the special case, where other features are set to mean observed value. Notice that the red line almost completely misses the local hill in the model structure. A single low dimensional slice of the mapping structure can easily miss prominent local interactions, when number of model dimensions is high.

A 2D-SA slice can explain a main effect and/or the possible interaction within two selected features. Figure \ref{partialFigure} only illustrates a 1D-SA slice projection, but represents the idea of any projection. The depicted model structure itself could infact be a 2D-SA projection of a higher dimensional model structure. Whether a given slice is a good generalization of the full mapping structure is unknown. A good generalization means that any parallel slices, where the fixed features are set to another combination, yield the same XYZ-visualization, with only perhaps a fixed offset in the prediction axis (Z) \cite{icebox}. We will for now term that such visualization has a high goodness-of-visualization. In section \ref{TheoDecompose} we will propose a metric for goodness-of-visualization. For a data structure with only additive effects and no interactions, the obtained model mapping structure is likely to have no interactions as well as any slice will be identical to its many parallel counterparts. In Figure \ref{partialFigure}, all the black parallel slices to the red slices give different projection lines in the mirror plane which could not be corrected by a simple offset. Therefore the model structure must have an interaction which cannot be seen in this projection alone. The iceBOX package displays multiple projection lines to search for masked interactions and is a good alternative to the forest floor approach \cite{icebox}.

A second concern is whether a given slice or slices extrapolate the training data. For a RF model with a satisfactory cross validated prediction performance, the mapping structure will represent the underlying data structure, but only within the proximity of the training data. Extrapolated areas of the mapping structure are far from guaranteed to represent an underlying data structure. Several different non-linear learners (RF, SVM, ANN, etc.) may easily have comparable model structures in the proximity to training data points, whereas far from the training set the models will heavily disagree. For RF models containing dominant interaction effects, the mapping structure on the borders of the training data becomes noise sensitive, as decision trees only can extrapolate parallel to feature axes, as the splits only are univariate. RF models only containing additive main effects have stable and smooth mapping structure at the borders of the training data. Model extrapolation of random forests with dominant interaction effects have been illustrated in supplementary materials.

SA plots remain a useful tool. When forest floor yield plots of similar structure, these plots generally represents the model mapping well. Visualization of multiple parallel projections, the so called ICE plots (individual conditional expectation) with the ICEbox package, can also reveal interactions. However multiple projection lines cannot directly filter out main effects by other features. These will tend to offset the projection lines on the prediction axis. Centered ICE (c-ICE) visualizations do adjust this offset by centering the prediction axis for all projections in one specific location \cite{icebox}.

A frequently used visualization method proposed by Friedman is the partial dependence plot (PD) which is the same as what Cortez and Embrechts later have termed data-based sensitivity analysis (DSA)\cite{Cortez2013,friedman2001}. In Figure \ref{partialFigure}, the green fat line in the mirror plane represents a partial dependence projection. Whereas 1D-SA and 2D-SA only project the slice intersecting e.g. the training data centroid, the partial dependence plot projects multiple slices. Each projected slice intersects one data point. The partial dependence line is the average prediction values of all slices. Thus, the obtained PD visualization summarizes all parallel slices of the mapping structure by averaging. To summerize, SA averages and then projects, whereas PD projects and then averages. ICE-plot projects many slices and do not aggregate. The PD approach may improve generalization across slices as it up-weighs the parts of mapping structure, that are well represented by data points. Still, interactions between varying and fixed features will be lost by averaging. Furthermore, the PD projections form a regular data grid spanned by the data observations. See the grid of black and green lines on the model structure surface in Figure \ref{partialFigure}. However, for data sets with high feature collinearity, data points will mainly be positioned in one diagonal of the grid, whereas the remaining part of the grid will span extrapolated parts of the model structure. This extrapolation occur for both SA, PD and ICE-plots.

Feature contributions was introduced by Kuz'min \cite{Kuz'min2011} for RF regression and elaborated by Palczewska \emph{et al} \cite{Palczewska2014} to also cover RF multi-classification. Feature contributions are RF predictions split into components by each feature. Feature contributions are essentially computed utilizing information from the tree networks of a RF model. Feature contributions have not before been used or understood in conjunction with the idea of function mapping structures. The contribution of this paper, is to show that feature contributions can be understood as a different way of slicing the mapping structure. From this insight the methodology, forest floor, was developed.

We have developed a number of tools to increase the usefulness of the forest floor methodology. These are: Out-of-bag cross validated feature contributions to increase robustness without increasing computation time, goodness-of-visualization tests to evaluate how well slices generalize the mapping structures and color gradients traversing mapping space to visually identify latent sources of interactions. Furthermore, the methods have been implemented as a freely available R-package, from which all mapping visualizations of this paper originate. The R-package forestFloor \cite{forestFloor2015} aims to assist the user visualizing a given RF model fit through a serious of appropriately chosen visualizations.

\section{Theory and calculation}
Here is provided a new notation for RF regression and classification to combine a mapping space representation with the feature contributions method developed by Kuz'min \cite{Kuz'min2011} and Palczewska \emph{et al.} \cite{Palczewska2014}. Moreover to obtain an exact decomposition of the model structure, we expand the previous notion of feature contribution to also cover the initial bootstrap and/or stratification step for each decision tree. For RF multi-classification we describe a probabilistic (K-1)-simplex prediction space, to improve the interpretation of feature contributions. Lastly we introduce how to calculate out-of-bag cross-validated feature contributions.

\subsection{Defining regression and classification mappings}
\label{TheoDefMap}
Any regression model $f_{r}$ can be seen as a mapping between a $d$-dimensional feature space $X \in \mathbb{R}^d$ and and a prediction scale $\hat{y} \in \mathbb{R}^1$
\begin{equation}
\label{regMap}
\hat{y} = f_{r}(X) \quad ,
\end{equation}
where $X$ represents the infinite set of points in the feature space. A subset of points in $X$ can be notated as e.g. $X_t$ where $t$ is a defined set. Single value entries of a countable subset of $X$ is notated as $x_{ij}$ where $i \in \{1,...,N\}$ ($N$ points) and $j \in \{1,...,d\}$ ($d$ features). $\hat{y}$ represents the entire prediction scale, where $\hat{y}_t$ could be a subset, if countable with point entries $\hat{y}_i$.

The entire mapping can be represented as a $d$-dimensional (hyper)surface $S$ in a $d+1$-dimensional mapping space $V$. $S$ can be understood as a learned model structure trained on a set of training observations, $t$. Obviously, if $d\in \{1,2\}$, then $S$ can conveniently be plotted by Cartesian axes as a 2D function plot or a 3D response surface (prediction as function of two features). Each label of a categorical feature can be assigned an integer value from 1 to \emph{K'} categories and thus also be plotted.

A classification model can be seen as a mapping from $X \in \mathbb{R}^d$ to $\hat{y} \in \{1,2,...,K\}$. Some models, as RF, provides a probabilistic prediction (pluralistic voting) of class membership $\hat{p}_{k}$ for any class $k \in \{1,2,...,K\}$ and assign the class membership hereafter. Thus, the probabilistic classification model $f_{c}$ is a mapping from $X$ to the probability space $P$, 
\begin{equation}
\label{probMap}
f_{c}(X) = P \quad .
\end{equation}

\begin{figure*}[htp]
\centering
\includegraphics{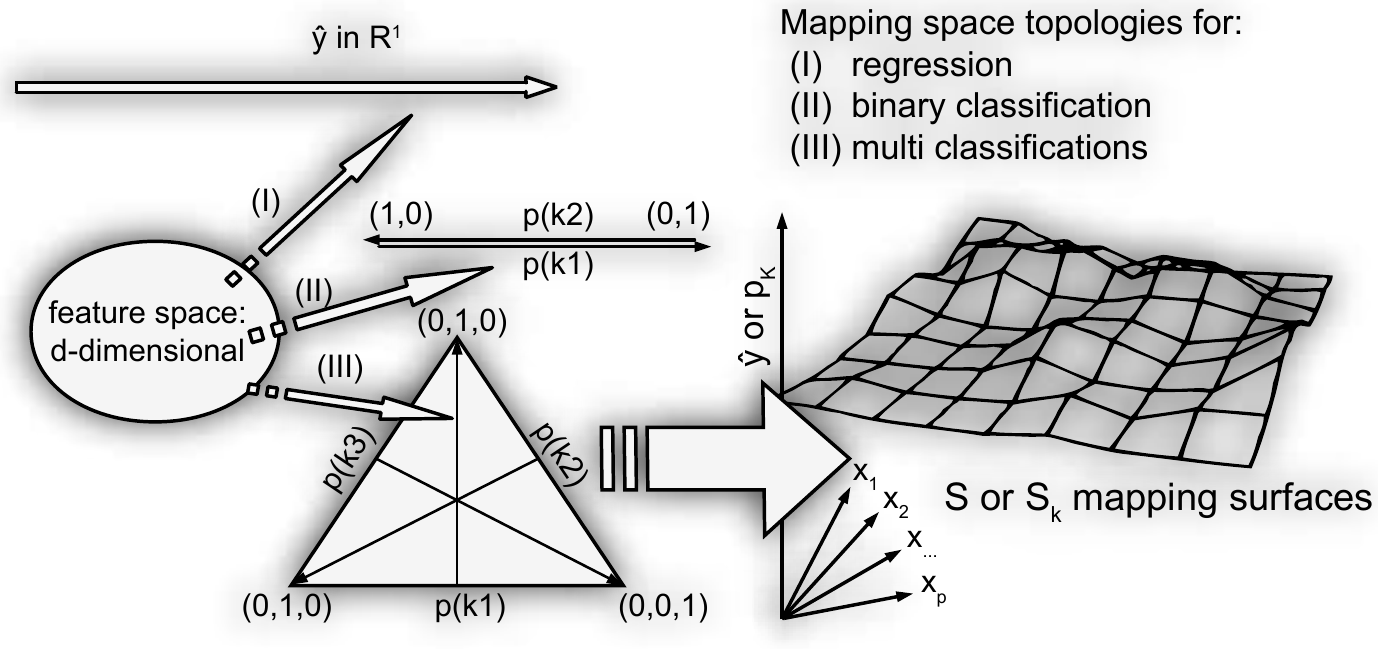}
\caption{Topologies of random forest model represented as a function mapping from $d$-dimensional feature space to one of the following prediction spaces: (a) regression, 1-dimensional scale; (b) binary classification, $K=2-1$ probability simplex reducible to a 1-dimensional probability scale; (c) multi-classification, probabilistic ($K-1$)-simplex. The mapping can be represented as a high-dimensional surface $S$, in a joined feature and prediction space linking any combination of features to a given prediction. For multi-classification, $S$ can be split into multiple $S_k$ surfaces describing predicted probability for each of $K$ individual classes.}
\label{mapTypes}
\end{figure*}

Any point in $P$ is a possible prediction $\hat{p}$ with a unique probability distribution over $K$ mutually exclusive classes, such that $\hat{p} = \{ \hat{p}_{1},\hat{p}_{2},...,\hat{p}_{K} \}$. As class memberships are mutually exclusive, the sum of the class probabilities is always one, $|\hat{p}|^1=1$. Therefore the probability space is a \emph{K-1} dimensional simplex \cite{OBrien2008}, which contains any possible combination of assigned probabilities to $K$ mutually exclusive classes, see Figure \ref{mapTypes} . The $K$ axes, which assign probability of 0 to 1, are not orthogonal, meaning it is not possible to modify the assigned probability of one class without affecting at least one other.

The classification mapping can be represented by simply joining the simplex-space with the feature space, but this would only allow a 2D or 3D visualization when $(d+K-1) \in \{2,3\}$, thus either maximally a 2 feature problem for 2 classes, or a 1 feature separation for 3 classes. Instead, this mapping can also be represented as $K$ separate $d$-dimensional surfaces $S_{k}$ in a $d+1$-dimensional space $V$ with $d$ axes representing features and one axis $\hat{p}_k$ representing the probability of either of the $K$ classes. Thus, we align the directions of all $K$ probability axes to reduce the dimensionality of the mapping space with $K-2$ dimensions. Then, any line parallel to the probability axis $\hat{p}_k$, will intersect every $S_{k}$ surface, describing the predicted probability of the $k^{th}$ class at this point of input features. The sum of predicted probabilities of all intersections for any such line will be equal to one. To summarize, multi classification model structures are more difficult to visualize, as each class adds another dimension to the mapping space. It is possible to plot the individual predicted probability of each class and overlay these plots. Figure \ref{mapTypes} summarizes the mapping topology for regression, for binary classification, and for multi classification.

RF mapping for both regression and classification can jointly be defined as
\begin{equation}
\label{RFmap}
\hat{y} = f(X) \quad .
\end{equation}

Here $\hat{y}$ is the \emph{c}-dimensional prediction space. For regression, $c=1$, $f$ maps to a 1-dimensional prediction scale. For classification, $c=K$ classes, and $f$ maps to a prediction vector space, where the $k^{th}$ dimension predicts the probability of class $k$. For classification the predictions $\hat{y}$ can be any point within the ($K-1$)-simplex. On the other hand, the training examples $y$ can only be of one class each, which are the $K$ vertices (corners) of the ($K-1$)-simplex.

We define a local increment vector, $L$, pointing from $\hat{y}_i$ to $\hat{y}_j$ in a prediction space of $c$ dimensions, such that

\begin{equation}
\label{LIdef}
L_{ij}=\hat{y}_j - \hat{y}_i =
\{\hat{y}_{j1}-\hat{y}_{i1},...,\hat{y}_{jc}-\hat{y}_{ic} \} \quad .
\end{equation}

For regression, where $(c=1)$, the local increment is a scalar with either a positive or negative direction. For classification, $(c>1)$, the local increment is a vector with $c$ elements, one for each class. Each node of a RF model fit is a prediction, which is a specific point in the prediction space. Local increments are the connections between nodes, describing the change of prediction. Computing the thousands or millions of local increments for trees and nodes, and sum these individually for each observation and feature is essentially the feature contributions method.

\subsection{Properties of random forest related to feature contributions}
\label{TheoRF}
\begin{figure*}[htp]
\centering
\includegraphics[width=1.00
\textwidth]{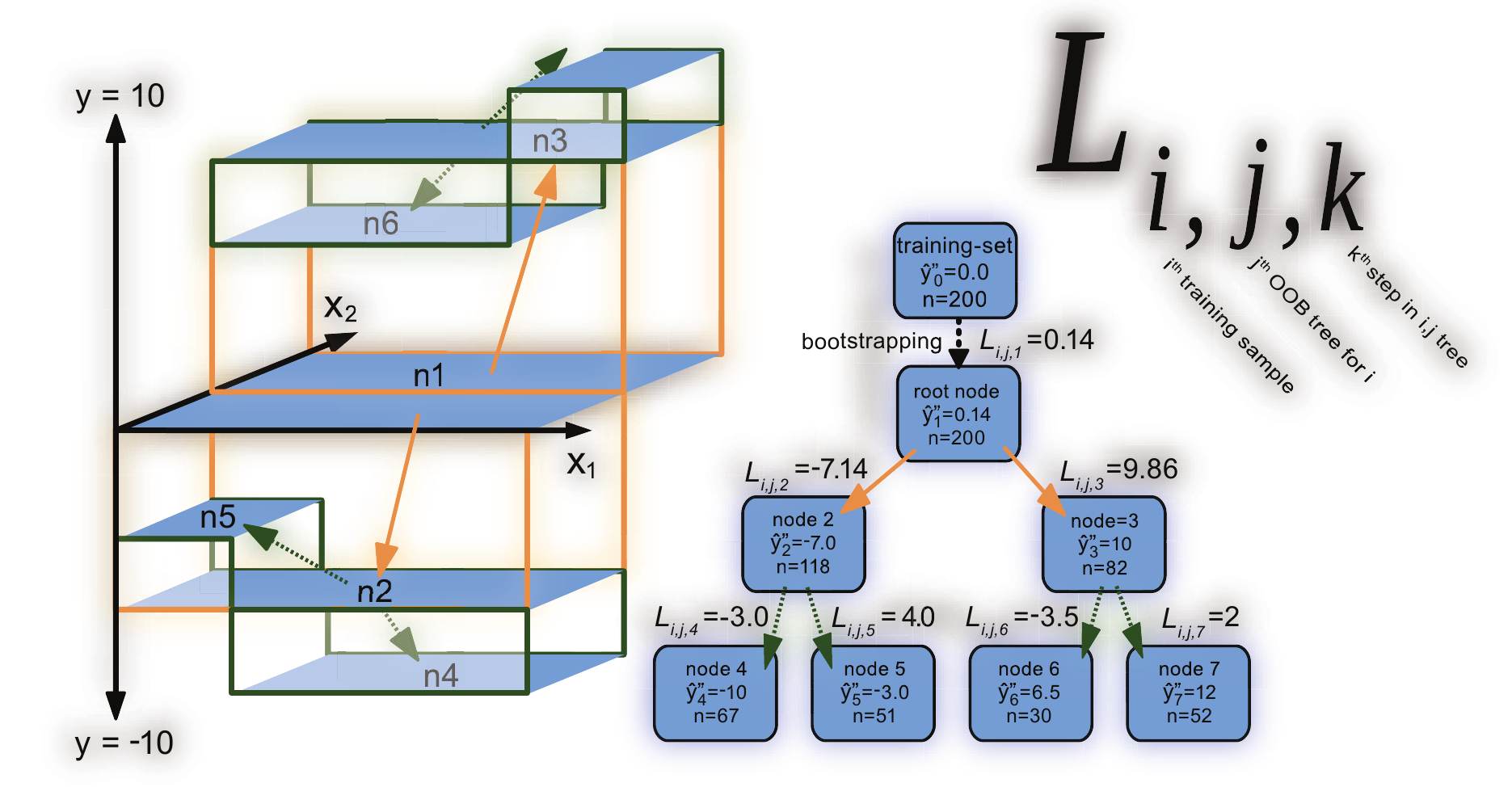}
\caption{Random forest and local increments explained. Left, an 3D illustration of a small regression tree of 7 nodes. Right, the same tree described by node means($\overline{u}$), node size($n$) and local increments $L_{ijk}$. $L$ is subsetted by observation, tree, node and feature. A observation falling in e.g.\ node 4, will have a prediction as the sum of the local increments in its path plus the grand mean of the training set.
}

\label{forestFigure}
\end{figure*}

RF is an ensemble of bootstrapped decision trees for either regression or classification. Figure \ref{forestFigure} illustrates how the RF algorithm operates for regression. For each of the trees (1 to $n_{tree}$) the training set is bootstrapped (random sampling with replacement). In average $ (\frac {N-1} N)^N \approx 0.37$ of $N$ observations will not be included in each bootstrap. These observations are called out-of-bag (OOB). Thus for any tree, a selection of observations will be 'in-bag' and used to train/grow the tree starting from the root node. Any node will have a node prediction which is defined by in-bag observations in that node.

\begin{equation}
\label{nodePredReg}
\hat{y}''_{j} = \frac{1}{n_j} \sum_{i=1}^{n_{j}} y_{ij}
\end{equation}

For a regression tree, the node prediction of the $j^{th}$ node $\hat{y}''_j$ is equal to the mean of in-bag target values in the $j^{th}$ node. Where $y_{ji}$ is the target value of the  $i^{th}$ observation in the $j^{th}$ node. $n_j$ is the number of observations in the $j^{th}$ node. Thus we are only computing a node prediction from in-bag elements.\par

For classification, the probabilistic node prediction $p_{jk}$ of the class $k$ of the node $j$ is equal to the number of in-bag observations of class $k$ divided with total number of in-bag observations in the node.

\begin{equation}
\label{nodePredClass}
\hat{p}_{jk} =\frac{n_{jk}}{n_j} \quad .
\end{equation}

A node prediction $\hat{y}''_{j}$ can also describe all class probabilities at once as a vector corresponding to a point in the $(K-1)$-simplex space.

\begin{equation}
\label{nodePredClass2}
\hat{y}''_{j} = \{ \hat{p}_{(j,1)},..., \hat{p}_{(j,K)} \}
\end{equation}

For classification $c>1$, the class probabilities of any node will always sum to 1 for any node:

\begin{equation}
\label{nodePredClassSumToOne}
|\hat{y}''_j|^1 = \sum_{k=1}^{K}{p_{jk}} = 1 \quad .
\end{equation}

Therefore, the elements of any local increment vector for classification, see Equation \ref{LIdef} will always sum to zero. This is not true for the local increment scalars of regression, $c=1$.


For an original RF implementation \cite{Liaw2002}, predictions of terminal nodes of classification trees are reduced to a single majority vote. Other implementations such as sklearn.randomForestClassifier \cite{sklearn} would rather pass on the probabilistic vote from terminals nodes and only on the ensemble level perform reduction by majority vote or just keep the full probabilistic average. In practice, implementations of feature contributions usually have to re-estimate node predictions. A feature contributions implementation such as forest floor should match the specific rule of terminal node predictions of the specific model algorithm.

A node is by default terminal if there are 5 or less in-bag observations left for regression or a single in-bag observation for classification. Any non-terminal node will be split into two daughter nodes to satisfy a loss-function. For regression the loss function is typically the sum of squared residuals.

For classification, a Gini criterion is used as the loss function. That is to select the split yielding the lowest node size weighted Gini impurity. Gini impurity ($g$) is 1 minus the sum of squared class prevalence ratios in nodes, $g = 1- \sum_{k=1}^K \hat{p}_{jk}^2$. Gini impurity is in fact the equation of a K-dimensional hypersphere, where $\sqrt{1-g}$ is the radius and all $\hat{p}_{jk}$ are the coordinates. The $(K-1)$-simplex space intersects this hypersphere where all prevalences sum to one, $1 = \sum_{k=1}^K \hat{p}_{jk}$. Therefore for a $K=3$ classification, a Gini loss function isobar appear as a 2D-circle, when visualized in the ($K-1$)-simplex space. One circular isobar is drawn in Figure \ref{LIclass}. The Gini loss function chooses the split placing two daughter nodes the furthest from the center of the ($K-1$)-simplex.

Splitting numerical features of ratio-, ordinal- or integer-scale is all the same for RF. A break point will direct observations lower or equal to the left node. Splitting by categorical features is to find the best binomial combination of categories designated for either daughter node. A feature with 8 categories will have $2^{8-1}-1 = 63$ possible binary splits. Any available break point are evaluated by the loss-function, but the RF algorithm is constrained to only access a random selection of the features in each node. The amount of features available, \emph{mtry}, can e.g.\ be a third of the total amount of features. This \emph{random variables subspace} and bootstrapping will ensure decorrelation of trees and feature regularization without overly increasing the bias of each fit. Each fully grown tree is most likely highly overfitted, as the individual predictions of each terminal node are dictated by 5 or less observations. Combining the votes of many overfitted but decorrelated trees form an ensemble with lowered variance and without increased bias. Out-of-bag(OOB) predictions are calculated for each terminal nodes. As OOB observations are not used actively in growing the trees of the forest, they can serve as an internal cross validation which yields similar results as a 5 fold cross validation \cite{Svetnik2003}. The prediction of individual trees are written as $\hat{y}'_{ij}$ for $i \in \{1,...,N\}$ observations predicted by $j \in \{1,...,n_{tree}\}$. The ensemble predictions are computed as

\begin{equation}
\label{pred}
\hat{y}_{i} = \frac{1}{n_{tree}} \sum_{j=1}^{n_{tree}} \hat{y}'_{ij} \quad ,
\end{equation}

and the OOB cross validated ensemble predictions $\tilde{y}_i$ are computed as
\begin{equation}
\label{OOBpred}
\tilde{y}_{i} = \frac{1}{|\tilde{J}_i|} \sum_{j \subseteq \tilde{J}_i}\hat{y}'_{ij} \quad ,
\end{equation}

where $\tilde{J}_i$ is the subset of $\{1,...,n_{tree}\}$ trees, where $i^{th}$ observation is OOB. $|\tilde{J}_i|$ is the size of the subset $\tilde{J}_i$. Thus let any training observation $i$ iterate through the $\tilde{J}_i$ subset of trees, defined as those trees where $i$ was not in-bag, and find the mean of terminal node predictions.

To obtain value/class predictions of new observations, the observations will be forwarded through all trees according to the established split rules. A tree prediction is dictated by the terminal node a given observation ends up in. The ensemble prediction of a RF model fit will by default be the average for regression and the majority vote for classification. Figure \ref{forestFigure} explains graphically the structure of a single regression tree by feature $x_1$ and $x_2$. First all bootstrapped observations exist within the node n1. The mean prediction value of n1 is in this example 0.14 a slight offset compared to the training set prediction mean of 0. The first split is over a break point in $x_2$, dividing n1 into n2 with low prediction value and n3 with a high prediction value. Both n2 and n3 are further split by $x_1$. Interestingly, n2 and n3 have almost opposite splits by $x_1$. In n2, high $x_1$ leads to a lower prediction, while reversely in n3. This illustrated tree have only grown 7 nodes. Nonetheless, the tree contains an interaction term, where high $x1$ only contribute positively to the prediction $\hat{y}$ when conditioned by high $x_2$.

\subsection{Local increments and feature contributions}
\label{TheoLI}
This section explains how feature contributions are computed. This paper expands the feature contributions defined by Palczewska \textit{et al} \cite{Palczewska2014} to also account for bootstrapping and/or stratification and to allow OOB cross validation. Feature contributions summarize the pathways any observation (a given combination of input features) will take through the many decision trees in a RF model. Each sub node of the trees holds a prediction, which is average observed target of observations populating it, see Equations \ref{nodePredReg} \& \ref{nodePredClass}. The sum of the many steps from node to node (local increments) is for regression exactly the resulting large step from the grand mean of the training set to the given numeric target prediction. Likewise for classification, the large step is from base rate to a probabilistic target prediction. A proof hereof is provided in supplementary materials. As these many small steps towards the final prediction is an additive process, it is possible to reorder the sequence of steps and end up by the same prediction.  The important implication hereof is that the RF model structure can be decomposed into additive sub models, each with the same dimensionality. As each sub model structure is the sum local increments of decision splits by one specific feature, each sub model structure tend to only describe the main effect of this one specific feature plus perhaps interactions with other features.

In order to efficiently describe how variations of feature contributions are computed, a notation of how to access any local increment in a given RF model fit is formulated. We define $L$ as a list of lists of lists containing all local increments. $L$ is defined in the following three levels (observations, trees, increments):
\begin{enumerate}
\item $L_i$ is a list with $i \in \{1,...,N\}$, and $N$ is the number of observations predicted by the forest. $i$ is the $i^{th}$ observation.

\item Each element of $L_i$, called $L_j$ is a list with $j \in \{1,...,n_{tree} \}$, and $n_{tree}$ is the number of trees in the ensemble. 

\item Each element of $L_j$, called $L_k$ is a list with $k \in \{1,...,n_{increment,i,j} \}$, and $n_{increment,i,j}$ is the number of increments encountered by the $i_{th}$ observation in the $j^{th}$ tree. 
\end{enumerate}

Note that $L$ can be ordered as a 2-dimensional array ($i$ observation, $j$ tree) where each element is a sequence of local increments specific for the $i^{th}$ observation in the $j^{th}$ tree. Overall, we can access any local increment in $L$ with $L_{ijk}$. Depending on the model type, $L$ will contain local increments as scalars for regression or as vectors for classification. The first local increment $k=1$ for any tree and observation in $L_{ijk}$ is the step from node 0 (training set) to node 1 (root node of tree). Thus the $k^{th}$ local increment steps from the parent node $k-1$ to a daughter node $k$. The local increment $L_{ijk}$ is the change of node prediction $\hat{y}''_{ijk}-\hat{y}''_{ij(k-1)}$

Equation \ref{LIeq2} describes how any prediction can be computed from $L_{ijk}$ as the sum of all local increments plus grand mean or base rate. A proof hereof can be found in the supplementary materials. 

The target prediction $\hat{y}_i$ is computed as

\begin{equation}
\label{LIeq2}
\hat{y}_{i} =  \frac{
\sum_{j=1}^{n_{tree}} \sum_{k=1}^{n_{increment,i,j}} L_{ijk}}
{ n_{trees}} 
+ \overline{y} \quad ,
\end{equation}

where $L_{ijk}$ is a local increment and where $\overline{y}$ is the grand mean or base-rate. The numerator is a scalar for regression and a vector for classification. The denominator,  $n_{tree}$, is always a scalar.

So far the prediction of the $i^{th}$ observation is the grand mean (regression) or the base-rate (classification) plus the sum of all local increments $L_{ijk}$ encountered by this $i^{th}$ observation divided by $n_{trees}$.

\begin{figure*}[htp]
\centering
\includegraphics{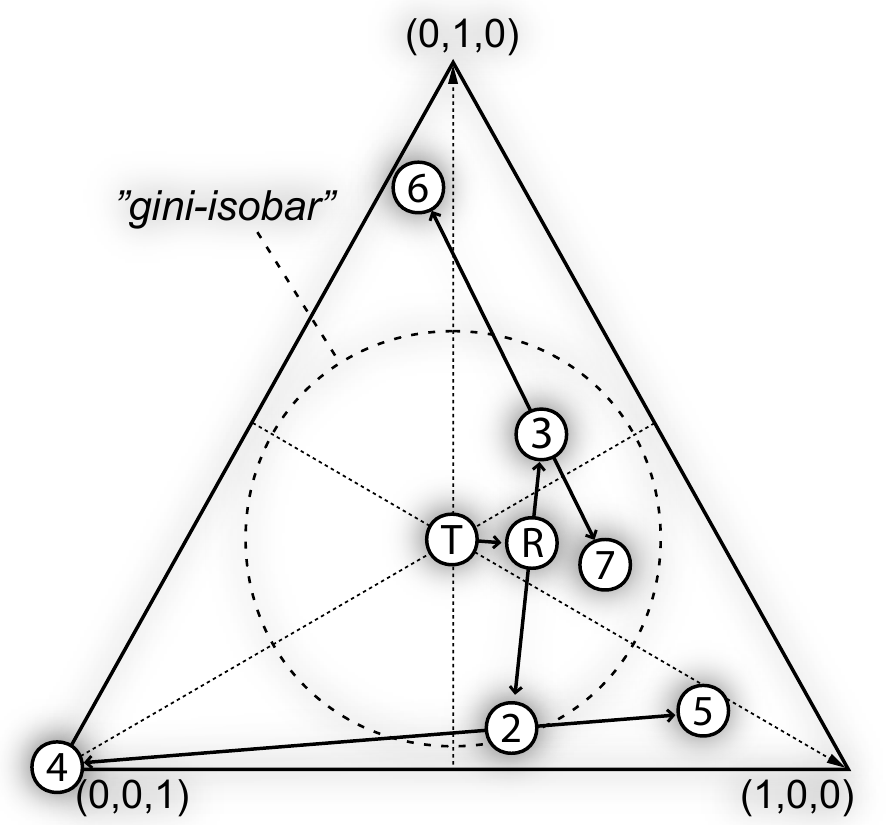}
\caption{A representation of how node predictions and local increments for a small classification tree with four terminal nodes. The first node in center represents the class distribution of a balanced training set (T). The bootstrap increment leads to the root node of the tree (R). The following local increments and nodes comprises the entire tree. Any split produces two local increments of opposite direction. The circle represents Gini loss function isobar. The further the two nodes (weighted by size) are from uniform class distribution the better a split according to the Gini loss function.} 
\label{LIclass}
\end{figure*}

Figure \ref{LIclass} is a new geometrical representation of local increments for a 3-class classification. Figure \ref{LIclass} is not intended as a model structure visualization, but rather as a representation of how decision trees branch out in the prediction space. Each node in the classification tree can be seen as a probabilistic prediction defining a point in a probabilistic ($K-1$)-simplex. Figure \ref{LIclass} depicts node predictions and local increments for a small tree with four terminal nodes. To this tree graph is appended a node (T) for training set to the root node of the tree. This train node represents the class distribution of the training set. The bootstrap increment leads to the root node. This step is often small and a result of random uniform sampling w/o replacement. If applying class stratification, the length and direction of this step can be controlled. Stratification corresponds to defining a prior expected class distribution, which will be the position of the root nodes in the prediction space. From here all trees will branch out from this point. The following local increments and nodes comprise the entire tree. Any split produces two nodes and two local increments of opposite direction. If not of equal node size, there will be one shorter local increment defined of many in-bag observations and one longer local increment defined of fewer in-bag observations. This is a consequence of that class distributions of daughter nodes multiplied by the node sizes and added together is exactly equal to class distribution of parent node multiplied by its node size. This symmetry effect can be found in Figure \ref{cmc.mainSimplex} in section \ref{resCMC}. For the unbalanced binary features \emph{wives' religion}, \emph{wives working} and \emph{media exposure} the prediction is offset a lot for a few observations, while the prediction of remaining many observations will only change a little in the exact opposite direction. For regression and binary classification such a direction is essentially one-dimensional and can be positive or negative. For multi classification the direction is a vector of K elements with the restriction that the sum of elements is zero. In Figure \ref{LIclass}, the circle represents a Gini loss function isobar. The further away (euclidean distance) nodes are placed from uniform class distribution the better a split according to RF Gini loss function. The best kind of split is one placing both daughter nodes onto two of the K vertices of the (K-1)-simplex. 
 
For the training set, a cross validated OOB-prediction $\tilde{y}$ can be formulated as

\begin{equation}
\label{LIeq3}
\tilde{y}_{i} =  \frac{
\sum_{j \subseteq \tilde{J}_i} \sum_{k=1}^{n_{increments,i,j}} L_{ijk}}
{ |\tilde{J}_i|}
+ \overline{y} \quad ,
\end{equation}

where $\tilde{J}_i$ is the subset of trees where $i^{th}$ sample is OOB. One can reason, that if Equation \ref{LIeq2} is true for any set of trees, then Equation \ref{LIeq3} must also be true for a given subset of any trees, such as the OOB subset $\tilde{J}_i$, see supplementary materials.

When predicting the training set with an RF model, any training observation $i \in \{1,...,N\}$ will have a high proximity to itself, that is, it will in any in-bag tree both define the in-bag node predictions of the terminal node and be predicted by the very same terminal node. For data sets with a high noise level this becomes a problem and the points $S_i$  of model structure $S$ will overfit the sampled training set observations $T_i$, and visualizations hereof will look more noisy. If the RF training parameter minimum terminal node size is increased and/or bootstrap sample size is lowered then training observation $i$ will have a lower influence on its own prediction and visualizations will not look noisy. \par

To compute feature contributions, the summed local increments over each observation and feature, it is necessary to keep a record of splitting features in each parent node. In Equation \ref{LIeq2}, the $i^{th}$ observation in the $j^{th}$ tree encountered the local increments for $k \in \{1,...,n_{increments,i,j}\}$. For this $i^{th}$ observation in $j^{th}$ tree, let $H_{ijl}$ be the subset of local increments where the parent node was split by the $l^{th}$ feature. The local increments of bootstrapping are assigned to \textit{feature 0}. The letter $H$ is used, as $K$ already is used to describe number of classes.

This distinction between OOB-predictions $\tilde{y}$ and regular test predictions $\hat{y}$ of training set now becomes important as how to feature contributions are defined. Previously \cite{Palczewska2014,Kuz'min2011} feature contributions have been defined for regression and classification analogous to this:

\begin{equation}
\label{FC}
F_{il} = \frac {
	\sum_{j=1}^{n_{tree}}
	\sum_{k \subseteq H_{ijl}}
	L_{ijk}
}{n_{tree}} \quad ,
\end{equation}

Here $F_{il}$, the feature contribution of the $i^{th}$ observation for the $l^{th}$ feature, is a subtotal of local increments $L_{ijk}$, where $k$ only iterates over $H_{ijl}$, which is those times the parent nodes were split by feature $l$.

This definition of feature contributions is fine if: (a) the noise level is low or (b) if feature contributions $F$ only is computed for some test set different from training set or (c) if the user is confident, that the model structure is not over fitted. It would be possible to cross validate by segregating the data set in a training set and test set to avoid over fitted visualizations. To discard data points is not desirable for a data set with limited observations. It would be possible to perform an n-fold cross validation, but n-fold random forests would be necessary to train.

We propose to compute feature contributions for the OOB cross validated predictions. OOB cross validated predictions are only the sum of local increments over trees where $i^{th}$ observation was OOB, see Equation \ref{LIeq3}. Analogously, we OOB feature contributions $\tilde{F}_{il}$ as

\begin{equation}
\label{FCoob}
\tilde{F}_{il} = \frac{
\sum_{j \subseteq \tilde{J}_i}
\sum_{k \subseteq H_{ijl}}
L_{ijk}
 } {|\tilde{J}_i|} \quad ,
\end{equation}

where $j$ only iterates the subset of trees $\tilde{J}_i$, and where $i^{th}$ observation was OOB. $|\tilde{J}_i|$ is the total number of times the $i^{th}$ observation was OOB and the size of the subset $\tilde{J}_i$. Equation \ref{FCoob} is used in forest floor visualizations to compute cross validated feature contributions of the training set predictions.

\subsection{Decomposing the mapping surface with feature contributions}
\label{TheoDecompose}
We can compute the OOB cross validated set of points $\tilde{S}_i = \{X_i,\tilde{y}_i\}$ for $i \in T$ the training set. That is the combination by training features $X_i$ and the cross validated predictions $\tilde{y}_i$, where $c=1$ for regression and $c>1$ for classification. To decompose $\tilde{S}_i$, then $\tilde{y_ i}\}$ is expanded with $\tilde{F}_{il}$, such that:

\begin{equation}
\label{FCsums2y}
\tilde{y_i} = \sum^{d}_{l=0} \tilde{F}_{il} + \overline{y} \quad .
\end{equation}

Likewise non cross-validated $\hat{y_i}$ is a sum of non cross-validated $F$,
\begin{equation}
\label{FCsums2yhat}
\hat{y}_i = \sum^{d}_{l=0} F_{il} + \overline{y} \quad .
\end{equation}

The ensemble prediction $\hat{y}$ or $\tilde{y}$ is equal to sum of local increments + grand mean / base rate, see Equation \ref{LIeq2},\ref{LIeq3}. As sequences of additive vectors can be rearranged, it is possible to compute sub totals of local increments of the full prediction. Feature contributions is just the subtotal of encountered local increments for the for the $i^{th}$ observation where the parent node was split by the $l^{th}$ feature.

Notice feature 0 ($l=0$) is included to accurately account for the normally small and negligible feature contribution of random bootstrapping. For an increasing number of trees, this bootstrapping feature contribution will approach zero. However, if the bootstrapping is stratified $F_{i0}$ and $\tilde{F}_{i0}$ is equal to local increment from training set base rate $\overline{y}$ to the chosen stratification rate in every root node.

\begin{figure*}[htp]
\centering
\includegraphics[width=1\textwidth]{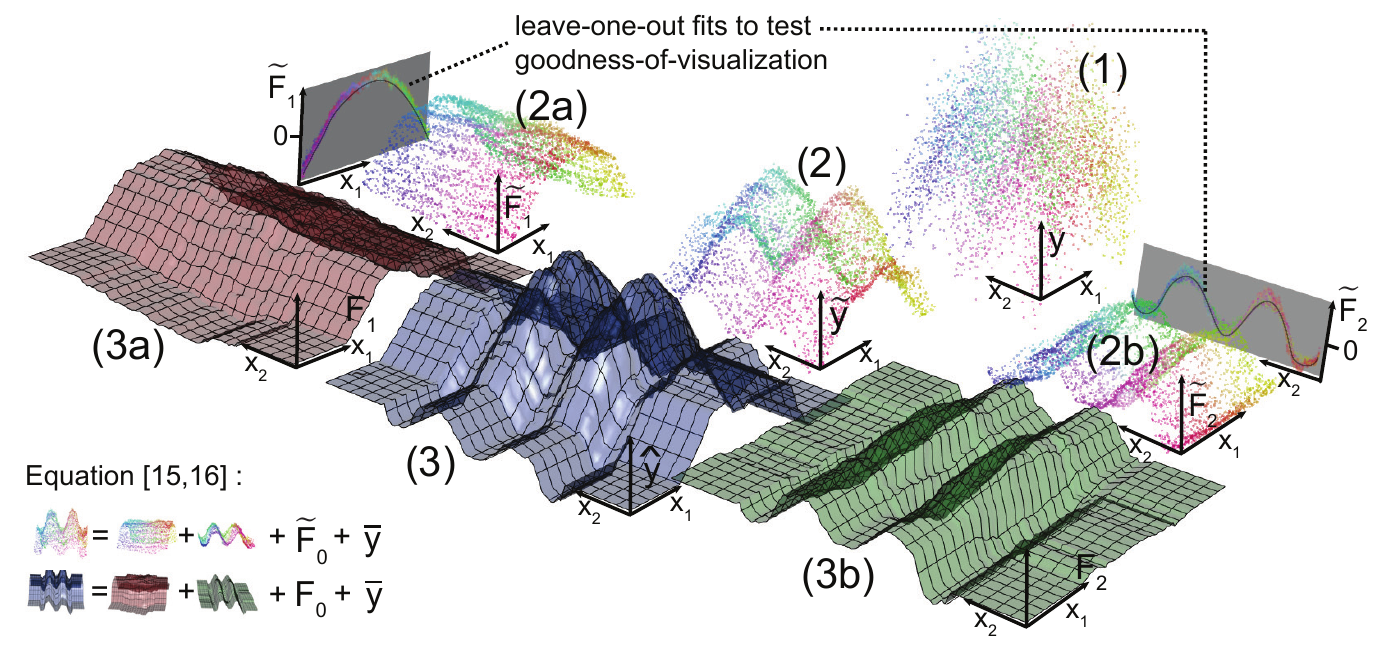}
\caption{(1) Simulated data set of 5000 observations,
$y_i = f(X_i) =  -(X_{i1})^2 - cos(X_{i2})) + \epsilon_i$ where $X_{i1}$ and $X_{i2}$ are drawn from a uniform distribution such that $X_1 \in [-\frac \pi 2 ;\frac \pi 2], X_2 \in [0;8 \pi]$. For all plotted points, a colour gradient (hue color wheel) is used to mark different combinations of $X_1$ and $X_2$. (2) Out-Of-Bag cross-validated predictions $\tilde{y}$ are plotted. (2a/2b) $\tilde{y}$ is decomposed into feature contributions $\tilde{F}_1$ and $\tilde{F}_2$ and projected into a 2D plane, see Equation \ref{FCoob} and \ref{FCsums2y}. Either contain almost only variance from the two main effects $-(X_1)^2$ or $cos(X_2)$. (3) Blue surface depict the full model structure, $\hat{y} = f(X)$. To either side (3a/3b) $\hat y$ is decomposed into $F_1$ and $F_2$, see Equation \ref{FC}. The sum of cross-validated feature contributions by each observation plus the grand mean $\overline{y}$ is equal to the cross-validated predictions, and vice versa for non-cross validated. $F_0$ is the corrections for random or stratified bootstrapping. If no stratification, $F_0$ will be negligibly small. This illustration also generalizes more input features/dimensions and probabilistic classification. 
}
\label{ffExplained}
\end{figure*}

Figure \ref{ffExplained} illustrates OOB cross validated feature contributions and regular feature contributions. A so called “one-way feature contribution plot” is a single feature contribution column plotted against the values of the corresponding feature. In Figure \ref{ffExplained} the "one-way feature contribution plot" can be seen as projections of $\tilde{F}$. Conveniently, the main effects of either feature $x_1$ and $x_2$ have been separated with feature contributions before the projection into the 2D plane. In Figure \ref{ffExplained}, the goodness-of-visualization fit to the projected feature contributions can be seen for both $\tilde{F}_{i1}$ and $\tilde{F}_{i2}$. If it is possible to re-estimate the set feature contributions e.g. $\tilde{F}_{i1}$ with some estimator $f$ only by the feature context of the visualization, it is certain, that no interactions have been missed. Thus the model structure do not contain any interaction effect with feature $x_1$. To quantify this we use a leave-one-out cross validation,

\begin{equation}
\label{GOV}
GOV(\hat{f}_\lambda) = cor(\hat{g_{.l}},\tilde{F}_{.l})^2 \quad ,
\end{equation}

here the goodness-of-visualization ($GOV$), is the pearson correlation between LOO predicted feature contributions. Where $\hat{g}_{il} = \hat{f}_{i}^{-i}(X_{i\lambda})$ is the leave-one-out prediction of the $\tilde{F}_{il}$ feature contribution of the $i^{th}$ observation for the $l^{th}$ feature. $\lambda$ is the features which are used to fit the estimator. When $\lambda = l$, $GOV$ quantifies how well feature contribution of the $l^{th}$ feature $\tilde{F}_{.l}$ is explained as a main effect. In Figure \ref{ffExplained} $\tilde{F}_{.1}$ is predicted by $X_{.1}$ and $\tilde{F}_{.2}$ is predicted by $X_{.2}$. GOV can also quantify other visualization contexts than main effect plots. E.g. in Figure \ref{regplot.3d} of result section the goodness of a visualization context of two features $x_3$ and $x_4$ is quantified, where $\lambda = \{3,4\}$.

\section{Materials and methods}

\subsection{Data and software}
\label{materialDataSoft}
The real datasets \emph{contraceptive method choice} (cmc) and \emph{white wine quality} (wwq) were acquired from the UCI machine learning repository \cite{Cortez2009,Lim1987}. All algorithms were implemented in R (3.2.4) \cite{R2015} and developed in Rstudio (0.99.892) \cite{Rstudio2015}. The main functionality is available as the R-package, \emph{forestFloor} (1.9.5) \cite{forestFloor2015}, published on the repository CRAN. If not stated otherwise all RF models was trained with the CRAN package \emph{randomForest} \cite{Liaw2002} by default parameters except keep.inbag=TRUE in order to reconstruct the individual pathways of observations through the trees. To reproduce result section, R scripts for each data example have been included in the package.

\subsection{Simulating toy data}
\label{materialToyData}
To demonstrate that the visualizations in the result section \ref{resStart} provide correct representations of the data structure, it is beneficial to use simulated (toy) data from a given hidden function. Such functions as Friedman\#1 and 'Mexican hat' are known examples \cite{Alhamdoosh2014}. To illustrate the principal functionality of forestFloor a new hidden function, $G$ is defined. $G$ is the ideal hidden structure, which cannot be observed directly. The toy function was defined as $G(X) + \epsilon = G^*(X) = y = x_1^2 + \frac 1 2 sin(2 \pi x_2) + x_3 x_4 + \epsilon k$ and was sampled 5000 times. $x_i$ were sampled from a uniform distribution $U(-1,1)$. The noise variable $\epsilon$ was sampled from a normal distribution $N(0,1)$ and $k$ was set such that the Pearson correlation $cor(G(X),G^*(X)) = 0.75$. Thus the true unexplainable variances component is roundly 25\% of the total variance. The level of detail, RF can capture from hidden structure G, declines as the noise increases.

\section{Results}
\label{resStart}
Three data sets were modeled with RF regression or RF classification and subsequently explored with forest floor. The examples demonstrate how feature contributions can be used to visualize the data structure and how to identify unaccounted interactions in a visualization.

\subsection{Random forest regression of \emph{toy data}}
\label{resToy}
\begin{figure*}[htp]
\includegraphics[width=1
\textwidth]{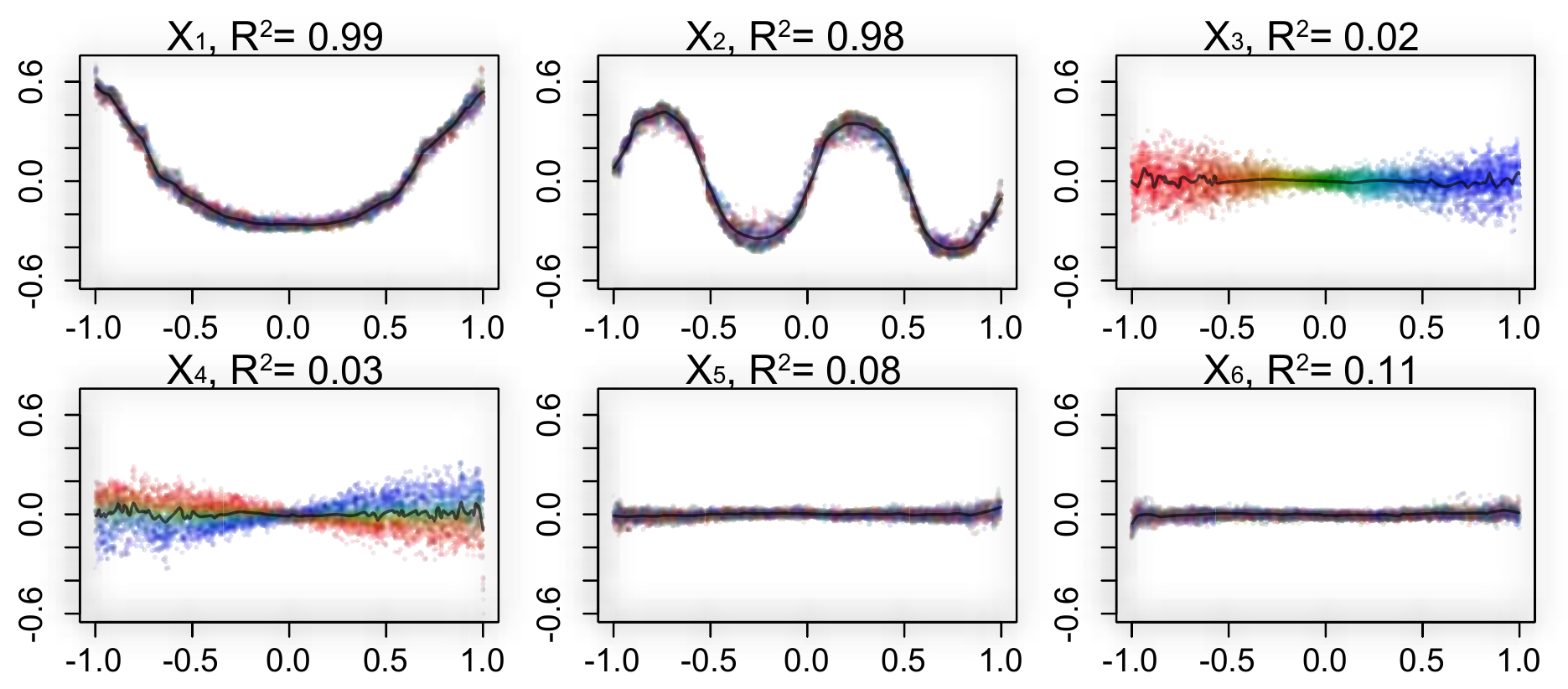}
\caption{Forest floor main effect plot of a RF mapping structure trained on hidden function $y={x_1}^2+\frac{1}{2}sin(\pi x_2) + x_3 x_4 + k \epsilon$. $x_5$ and $x_6$ have no relation to $y$ and were included only to illustrate a base line signal. A color gradient parallel to $x3$ is applied to identify latent interaction with $x4$. Leave-one-out k-nearest neighbor gaussian kernel estimation provides goodness-of-visualization(black line \& $R^2$ correlation) to evaluate how well each feature contribution can be explained as a main effect.}
\label{regplot.main}
\end{figure*}

A default RF regression model was trained on the toy data set with a hidden structure, $y = x_1^2 + \frac 1 2 sin(2 \pi x_2) + x_3 x_4$. Figure \ref{regplot.main} plots feature contribution of all six features against the training set feature values of the toy data. This type of plotting illustrates the main-effects, as feature contributions by each feature were plotted against their respective feature values. Hereby, the mapping surface $S$ was visualized as the sum of $d$ partial functions(black-lines), one for each feature. As the feature contributions retained any variance (main effects + interactions) associated with the node splits by each feature, it was possible to visually verify and test the goodness-of-visualization. Notice that main effect plots of $x_1$ and $x_2$ form nonlinear patterns representing the underlying additive $x^2_1$ and $\frac 1 2 sin(2 \pi x_2)$ contributions to the target y. Therefore, the leave-one-out $R^2$ goodness-of-visualization was $>0.95$ for both these plots. As the explained variance of feature contributions of $x_1$ and $x_2$ was more than 95\% when fitted as main effects, there was no considerable unaccounted interactions. On the other hand, feature contributions of $x_3$ and $x_4$ were poorly explained in main the effect plots. The GOV was poor, less than $R^2<0.1$. It was hence concluded that plotting the one-way feature contributions of $x_3$ and $x_4$ did not assist to explain the structure of $S$. Feature contributions of $x_5$ and $x_6$ were also poorly explained but contained no large variance and were therefore not interesting to explore further. The features $x_5$ and $x_6$ could also be identified as unrelated to the target $y$ for having a very low variable importance (not shown). To include such uncorrelated/unrelated features illustrated the base line of random fluctuations in the mapping structure. This helped to assess whether a given local structure only was a random ripple.

\begin{figure*}[htp]
\centering
\includegraphics{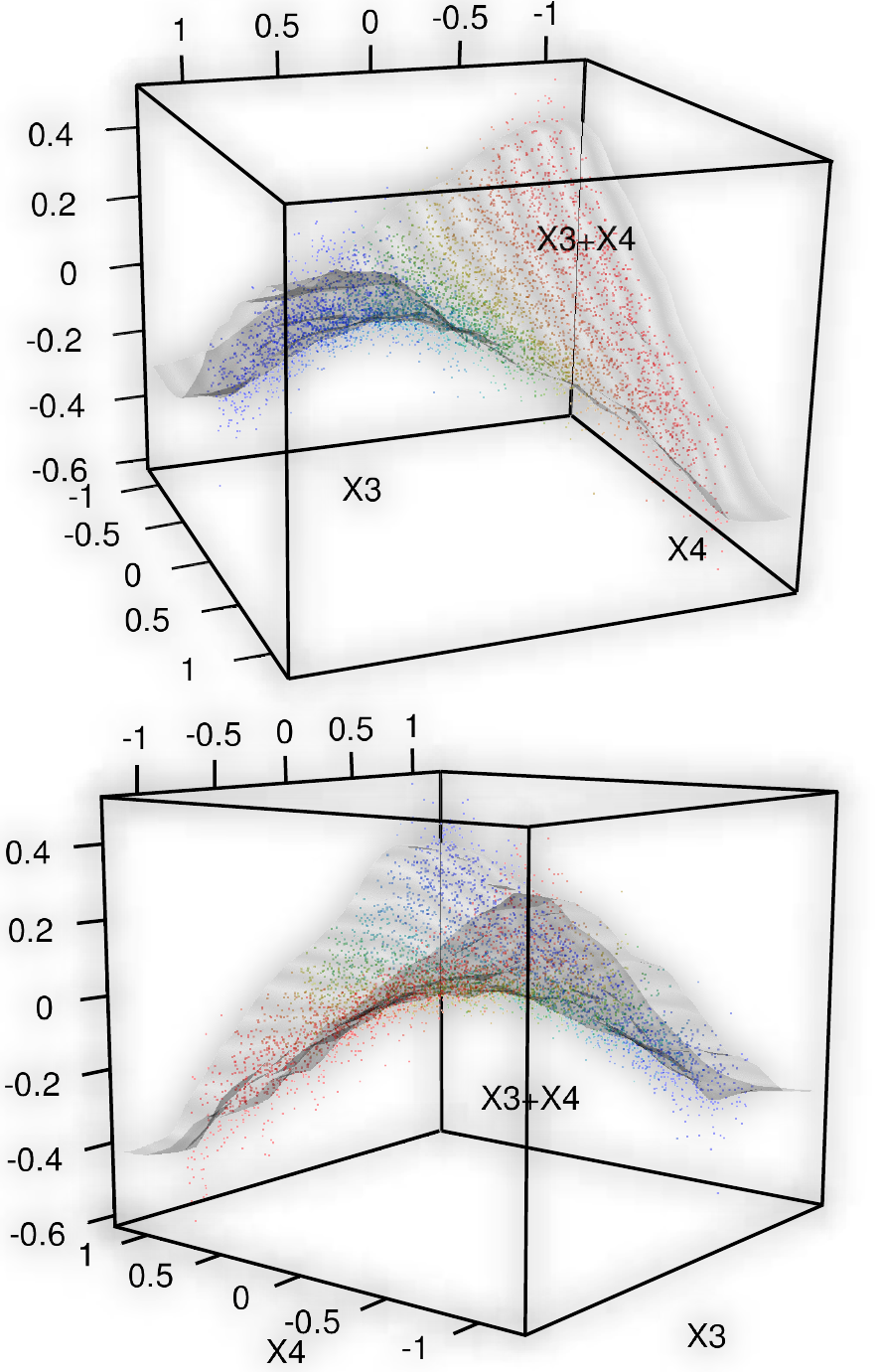}
\caption{One forest floor interaction plot. XY-plan represent feature values $x_3$ and $x_4$ and Z-axis is the summed feature contributions of $\tilde{F}_{i3} + \tilde{F}_{i4}$. goodness-of-visualization is evaluated with leave-one-out k-nearest neighbor gaussian kernel estimation (grey surface, $R^2= .90$). This indicates no remaining latent interactions related to features $x_3$ and $x_4$.}
\label{regplot.3d}
\end{figure*}

As the feature contributions of $x_3$ and $x_4$ were inadequately accounted for, a broader context was needed to understand the hidden structure. To identify interactions relevant for the feature contribution of $x_3$ a color gradient (red-green-blue) was applied in mapping space $V$ along the $x_3$ axis. The color of any other observation in any other plot was decided by its projected position on the $x_3$ axis. Low values were assigned red and high values blue. Figure \ref{regplot.main} depicts the main effects feature contribution plot of $x_1$,...,$x_6$ with the applied color gradient to $x_3$. Any main effect feature contribution plot of features who neither correlate and neither interact with $x_3$ will show a random color pattern. Such features were $x1$, $x2$, $x5$ and $x6$, which neither correlated nor interacted with $x_3$. Plots of only correlated features would reproduce the same horizontal color pattern. In the extreme case, a feature identical to $x3$ would reproduce the exact same horizontal color pattern. Plots of only interacting features would reproduce the color gradient vertically along the feature contribution axis. A combination of correlation and interaction would make the color gradient reappear diagonally. In Figure \ref{regplot.main} the color gradient suggests, that $x_3$ interacted with $x_4$ due to the vertical color gradient in the plot of $x_4$. In Figure \ref{regplot.3d} their combined feature contributions were plotted in the context of both feature $x_3$ and $x_4$. In this 3D plot it was observed, that the 2D rule of color gradients of interacting features was a basic consequence of perspective. Both color patterns of $x_3$ and $x_4$ could be reproduced by rotating the 3D plot. In this 3D plot, there was no large deviation of feature contributions from the fitted grey. Thus, it was evident that any structure of $S$ related to $x_3$ and $x_4$ were well explained in the joined context of both features $x_3$ and $x_4$.
The GOV of this fit was $R^2>.9$. Therefore, this second order effect plot was an appropriate representation of how $x_3$ and $x_4$ contribute to the target $y$. The depicted saddle-point structure of Figure \ref{regplot.3d} was expected, as the product of $x_3$ and $x_4$ contributed additively to the target $y$. Overall, the model surface $S$, could be represented by two one-way plot of $x_1$ and $x_2$ and one two-way plot of $x_3$ and $x_4$. Hereby the hidden structure of the toy data was fully recovered.

\subsection{Random forest regression of \emph{white wine quality} (wwq)}
\begin{figure*}[htp]
\centering
\includegraphics[width=1
\textwidth]{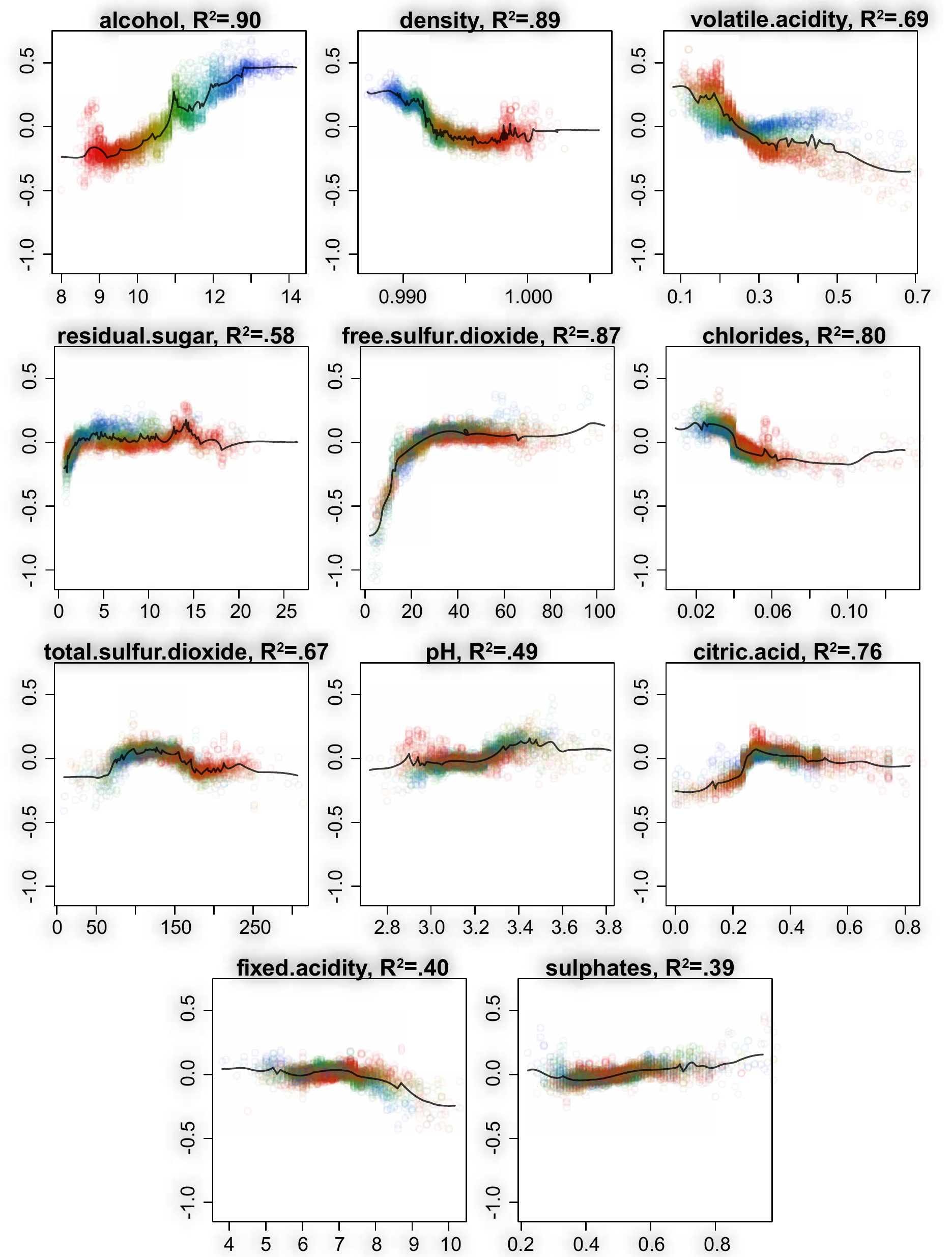}
\caption{Forest floor main effect plots of random forest mapping structure of model predicting panel ratings of 4900 white wines on basis of chemical properties. The plots are arranged according to variable importance. X-axis are variable values and Y-axis the corresponding cross validated feature contributions. Color gradient in all plots are parallel to the feature \emph{alcohol} (content w/w). goodness-of-visualization is evaluated with leave-one-out k-nearest neighbor estimation (black line , $R^2 values$)}
\label{wwq.1.main}
\end{figure*}

\begin{figure*}[htp]
\centering
\includegraphics{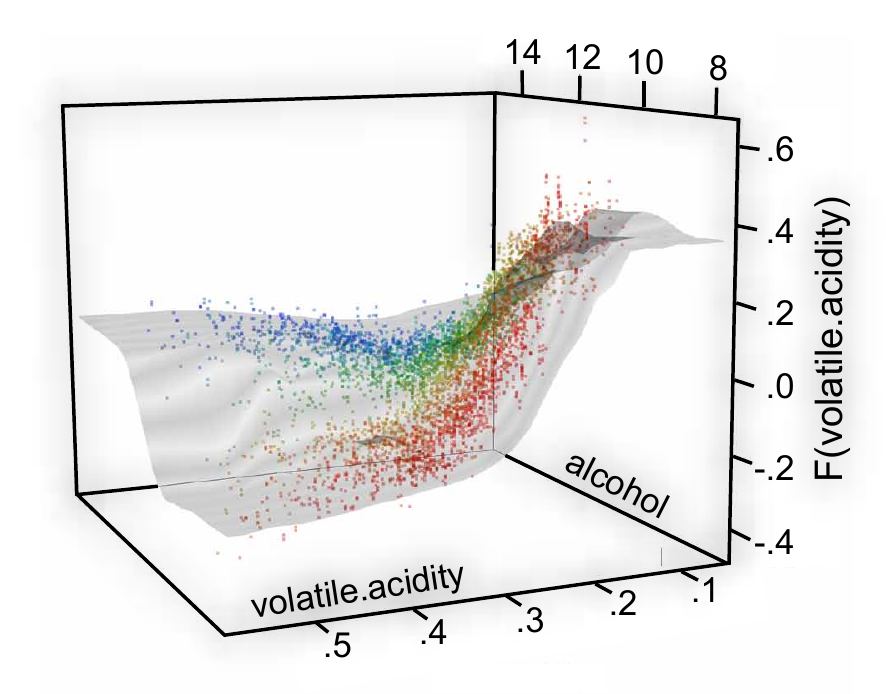}
\caption{Forest floor interaction plot: Feature contribution of \emph{volatile acidity} versus feature values of \emph{volatile acidity} and \emph{alcohol}. Color gradient is parallel to \emph{alcohol} axis. goodness-of-visualization is evaluated with leave-one-out  k-nearest neighbor estimation (grey surface and $R^2 = 0.93$)}
\label{wwq.volatile2way}
\end{figure*}

The previous example of forest floor visualization was an idealized example with uncorrelated features and either representing clear main effect or clear interaction effects. The white wine quality data set (wwq) is an example of mixed main effects and interactions by most features. The target, consumer panel ratings(1-10) of wines, was predicted on basis of 11 chemical features. A default RF model was trained and explained 56\% of variance and the mean absolute error was 0.42 rating levels matching the previous best found model performance \cite{Cortez2013}. To explore the model structure of $S$, first all main effect plots were inspected. Figure \ref{wwq.1.main} depicts all plots by all 11 features. Features were sorted in reading direction by variable importance to present most influential feature first. A color gradient along the most influential feature, \emph{alcohol}, was applied to search for interactions. Hereby it was observed that \emph{density} was negatively correlated with \emph{alcohol}, that \emph{volatile acidity} interacted with \emph{alcohol} and that \emph{residual sugar} both correlated and interacted with \emph{alcohol}. The observed correlation between \emph{residual sugar}, \emph{density} and \emph{alcohol} is trivial, where low-density \emph{alcohol} linearly lowers \emph{density} while high-density \emph{residual sugar} increases \emph{density}. Close to 98\% of the scaled variance of these three features can be described by two principal components. This information redundancy was expected to affect variable importance of the three implicated features and to lower the general variance of the respective feature contributions. Although the overall structure suggested that alcohol content in general was associated with higher preference scores, there was a local cluster identified as low \emph{alcohol}, high \emph{residual sugar} and low \emph{pH} which was associated with high preference scores also. Figure \ref{wwq.1.main} suggested that wines could achieve a high preference score when \emph{residual sugar}$\approx$17, \emph{pH}$\approx$2.9, \emph{citric acid}$\approx$.35 and \emph{fixed acidity}$<$7 despite a low alcohol content. Such white wines was perhaps by the consumer panel attributed fruity and fresh. Any found interaction could be investigated with several color gradients and two-way forest floor plots. It was chosen to investigate the interactions of \emph{volatile acidity}, as this feature was the third most important feature, whereas the goodness-of-visualization of the one-way forest floor plot was only $R^2=0.69$. Two-way forest floor plot was therefore a more suitable representations of this effect. The color gradient along alcohol content already suggested a notable interaction between \emph{volatile acidity} and \emph{alcohol}. Figure \ref{wwq.volatile2way} depicts the two-way forest floor plot of feature contributions of \emph{volatile acidity} in the context of itself and the feature \emph{alcohol}. The goodness-of-visualization was then $R^2=0.94$. Therefore, the residual variance of feature contributions not explained by this plot was low.
For wines with alcohol content more than 10\% (blue area) \emph{volatile acidity} appeared slightly positively to preference score. For wines with lower than 10\% \emph{alcohol} (red area) \emph{volatile acidity} appeared to contribute negatively to preference score.

\subsection{Random forest multi-classification: \emph{Contraceptive method choice} (cmc)}
\label{resCMC}
To illustrate the capabilities of forest floor for multi-classification the data set cmc was chosen. The data set originates from a survey of 1473 non-pregnant wives in Indonesia in 1987 comparing current choice of contraception with socioeconomic features. These features were, \emph{wives' age} (16-49), \emph{wives' education level} (1-4), \emph{husbands' education} (1-4) , \emph{n\textunderscore children} (0-16), \emph{wives' religion} (0 (not islam), 1 (islam) ), \emph{wives working} (0 (yes), 1(no)), \emph{husbands' occupation} (I,II,III,VI), \emph{standard-of-living index} (1-4), \emph{media exposure} (0=Good, 1=not good) and the target \emph{contraceptive method choice} (1=no-use (629), 2=long term(333), 3=short term (511)).

In the \emph{cmc} data set the choice of contraception was far from fully described by the available features \cite{Lim2000}. The OOB cross validated RF model error-rate was $.44$.  Assuming wives did not use contraception (the most prevalent case) yielded a $\frac{629}{1473} =.57$ error rate. Anyhow, if the RF model performance would be regarded as good by domain specialists, the model structure could possibly provide insights to the socioeconomic mechanisms in play. Hyper parameters \emph{Sample size} and \emph{mtry} were tuned to yield the best OOB cross validated performance. Optimal parameters was found to be bootstrap \emph{sample size}= 100 and \emph{mtry} = 2. A lower \emph{sample size} can increase robustness by tree decorrelation but also introduce more bias. To lower \emph{sample size} of trees can be advantageous, when explained variance component is less than 50\%. Thus a RF model different from default settings, was chosen to slightly improve predictions and to simplify/smooth the mapping structure to explore. Hereby the mapping structure may better represent the underlying social/economic mechanisms, that the specific data structure of survey reflects.

\begin{figure*}[htp]
\centering
\includegraphics[width=1
\textwidth]{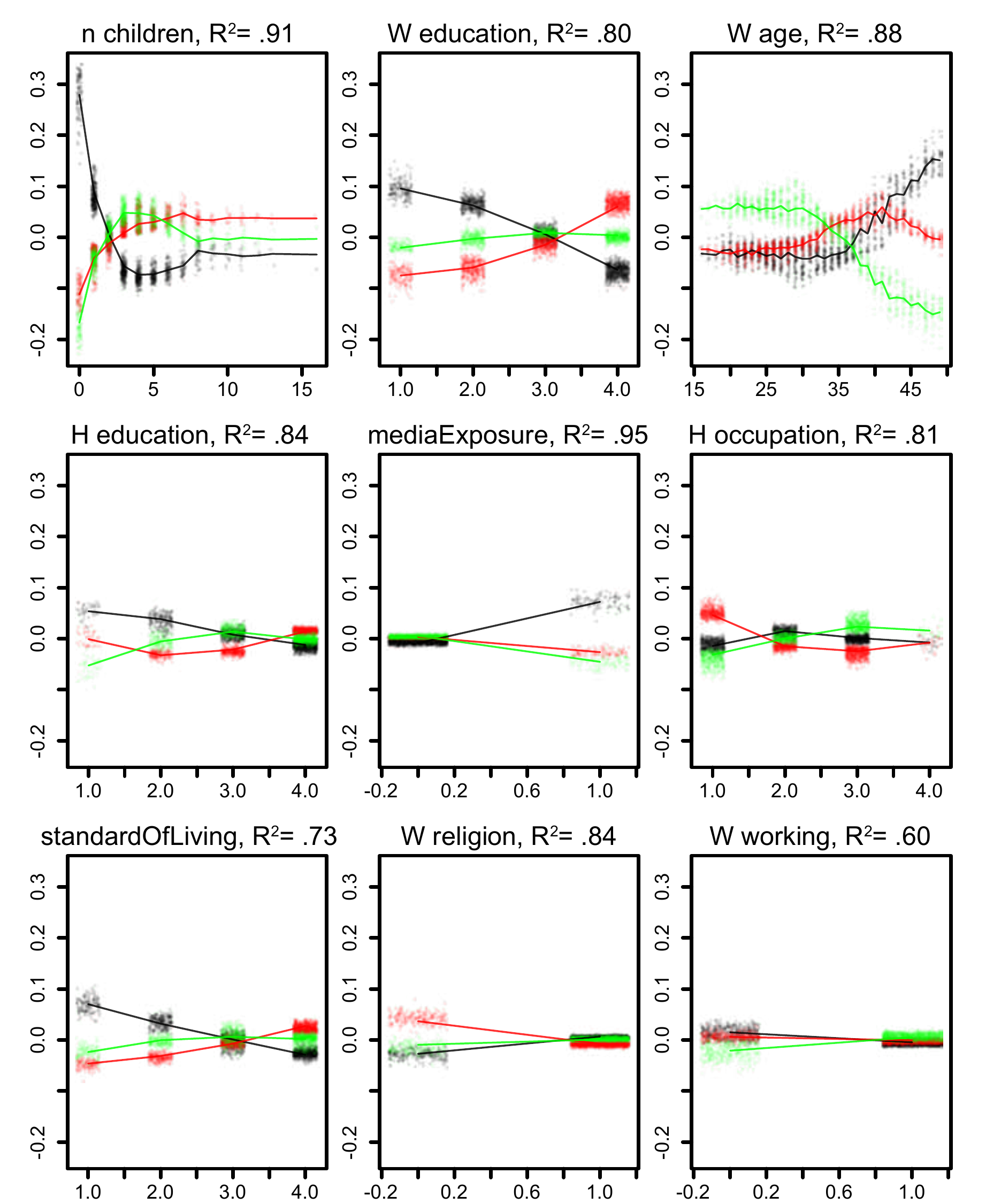}
\caption{Cross validated feature contributions for each feature for each class(black, red, green) and for all training observations plotted against the corresponding feature values. Categorical features are coded with integers. Feature contributions can be understood as change of predicted class probability attributed to a given feature.}
\label{cmc.mainAlign}
\end{figure*}

Three types of plots were constructed to investigate the mapping structure. As the number of features was $d=9$ and number of classes was $c=3$, a full dimensional mapping space visualization would require 12 dimensions. As shown in Figure \ref{mapTypes}, probability axes can be aligned along the y-axis, to reduce the number of dimensions to represent prediction space to only one. Also, when the cross validated predictions were decomposed into cross validated feature contributions, only 2 dimensions were needed to plot any main-effect. These plots resembled one-way forest floor regression plots although coloring was reserved to identify class of predicted probability. Otherwise each class by each feature would need to be plotted separately. Black assigns no usage. Red assigns long-term usage and green assigns short-term usage. Figure \ref{cmc.mainAlign} illustrated the main effects of each feature of a RF-fit, the y-axis describes the additive change of predicted probability for each observation for each each class. The actual feature value for each observation was depicted by the x-axis. Thus any observation were placed three times in each plot by the same feature value in three colors once for each three classes. The sum of changed probability over classes for any observation must be zero, see Equation \ref{nodePredClassSumToOne}. Overall, Figure \ref{cmc.mainAlign} showed that main effects were dominant, as most variance was explained by the respective features. \emph{n\textunderscore children} was the most important feature strongly predicting (probability change up to +/- .30) that wives with 0 or 1 child tended not to use contraception. On the other hand, more than 4 children predicted a slight increase in either type of contraception. Except for a preference separation for long-term contraception over short-term for wives with more 7 children, the \emph{n\textunderscore children} feature was not found useful to predict the choosing  betwen the two types of contraception. \emph{Wives's education} especially separated between no-use of contraception and long-term use, where lowest level predicted up to +/-10\% probability change. With more education the wives tended to use long-term contraception over no usage. The use of short-term contraception was comparably unchanged as a function of \emph{wives' education}. \emph{Wives' age}, the third  most important feature, favored short-term contraception for wives younger than 30, while long-term and no contraception for wives elder than 30. After 40 years, either use of contraception declined. \emph{Husbands' education} elicited same pattern as \emph{wives' education} though size of effect was half. A small subgroup of 7\% was reported to have a not good \emph{media exposure} and this predicted a probability increase in no contraception of 8\%. Type of \emph{Husband' occupation} favored for category \emph{I} long-term by 5\%  over short-term, whereas category \emph{III} predicted an opposite 3\% effect. Standard of living predicted a pattern much similar to \emph{husband's eduction}. A small subgroup (15\%) of wives were not muslim, and this predicted a 5\% increase in short-term contraception over long-term usage and no usage. Lastly for a subgroup of 25\% working wives was predicted a very slight increase (2\%) of no-usage over short-term. 

\begin{figure*}[htp]
\centering
\includegraphics[width=1
\textwidth]{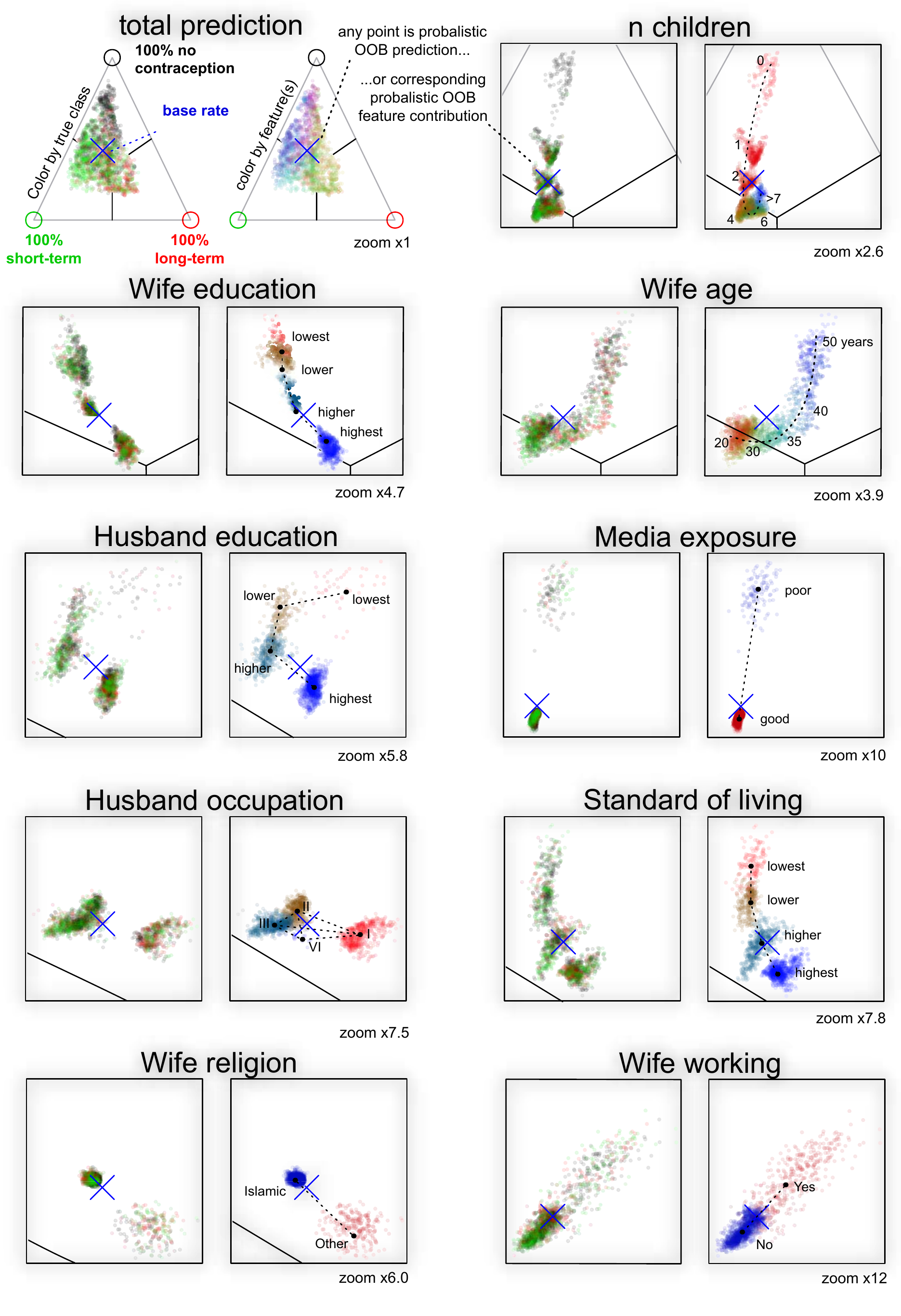}
\caption{From top left: Cross validated predicted class probability colored by true class and a PCA color gradient describing observation diversity. Following pairs of plots, were the predicted probability decomposed into feature contributions. Left colored by true class, right colored by corresponding feature value. Red is minimal value, blue is maximal value. Blue cross is class base rate of training set. Dashed lines are drawn manually to assist interpretation of main effects.}
\label{cmc.mainSimplex}
\end{figure*}

The main effects for this 3-class problem could also be depicted as a series of $(3-1)$-dimensional simplexes, where the position in the triangle depicts the predicted probability distribution for any observation. Colors can either depict true class (black: no-usage, red: long-term and green: short term) or colors can depict a feature (low value (red), middle (green), high(blue)). Figure \ref{cmc.mainSimplex} depicts all main effects in bi-simplex plots, with left simplex colored by cross-validated true class separation, and right simplex colored by feature value distribution across the simplex space. Figure \ref{cmc.mainSimplex} depicts 10 pairs of simplexes. Lines were added to the simplexes to illustrate majority vote. Only 17\% of wives were predicted to use long-term contraception even though 22\% of the sample population did so. Because RF models effectively used the sampled base rate as prior (marked as a blue cross) and the effective separation was weak, predictions tended to be skewed towards largest class away from smallest class. A different prior than the sampled base rate could be set by stratified bootstrapping of each tree in a random forest model. E.g.\ to stratify sampling by target class would move the blue cross to the middle of the simplex, and roughly a third of predictions would fall into either class. Stratified bootstrapping would e.g. be reasonable if the preferred contraception is expected to be different in the full population than in the training population.

In the second total separation simplex, to present an overview of any differences in socioeconomic status, principal component analysis was used to reduce the full feature space to two principal color components. Here a purple cluster indicated no-usage, a green cluster was shifted towards long-term usage, light blue cluster predicted short-term usage, and a dark-blue cluster predicted short-term or no usage. The color separation was not perfect, partly because the separation problem was difficult and partly because PCA cannot fully characterizes a potential nonlinear mapping surface of randomforest. To colour be several features at the same time, seemed to be most useful for data sets with high linear feature collinearity.

The left of following bi-plots of simplexes depicted the effective separation of true class separation by any feature contribution. The right simplex depicted the separation as a function of the corresponding feature (by color). This second simplex could be used both to illustrate the main effect of each feature and to assess whether higher order effects were present. For features with small set of levels such as womans education, a separation in four clusters (red(1), brown(2), pale blue(3), deep blue(4)) could be seen. Education level 1 and 2 were partly joined. The local centroids of these cluster was interpreted as the main effect, and the deviation from the centroids as higher order effects + unfiltered noise. For all simplexes the global centroid and prior is the (blue cross).

The series of bi-plot simplexes of Figure \ref{cmc.mainSimplex} could illustrate with finer detail the predicted probability distribution for any observation, whereas the precise feature value was depicted with less fidelity than in Figure \ref{cmc.mainAlign}.

The three features \emph{media exposure}, \emph{wives' religion} and \emph{wives working} were binary and showed the largest change of predicted probability in the smallest subgroups. This observation was regarded trivial, as the group size weighted probability change across a binary feature split must have equal size. Thus few observations can change prediction a lot, if many observations only change prediction a little in a opposite direction. This was regarded a property for all binary decision tree models and Figure \ref{LIclass} in Section \ref{TheoLI} depicted a similar pattern of how local increments would propagate in a probability simplex.

\begin{figure*}[htp]
\centering
\includegraphics{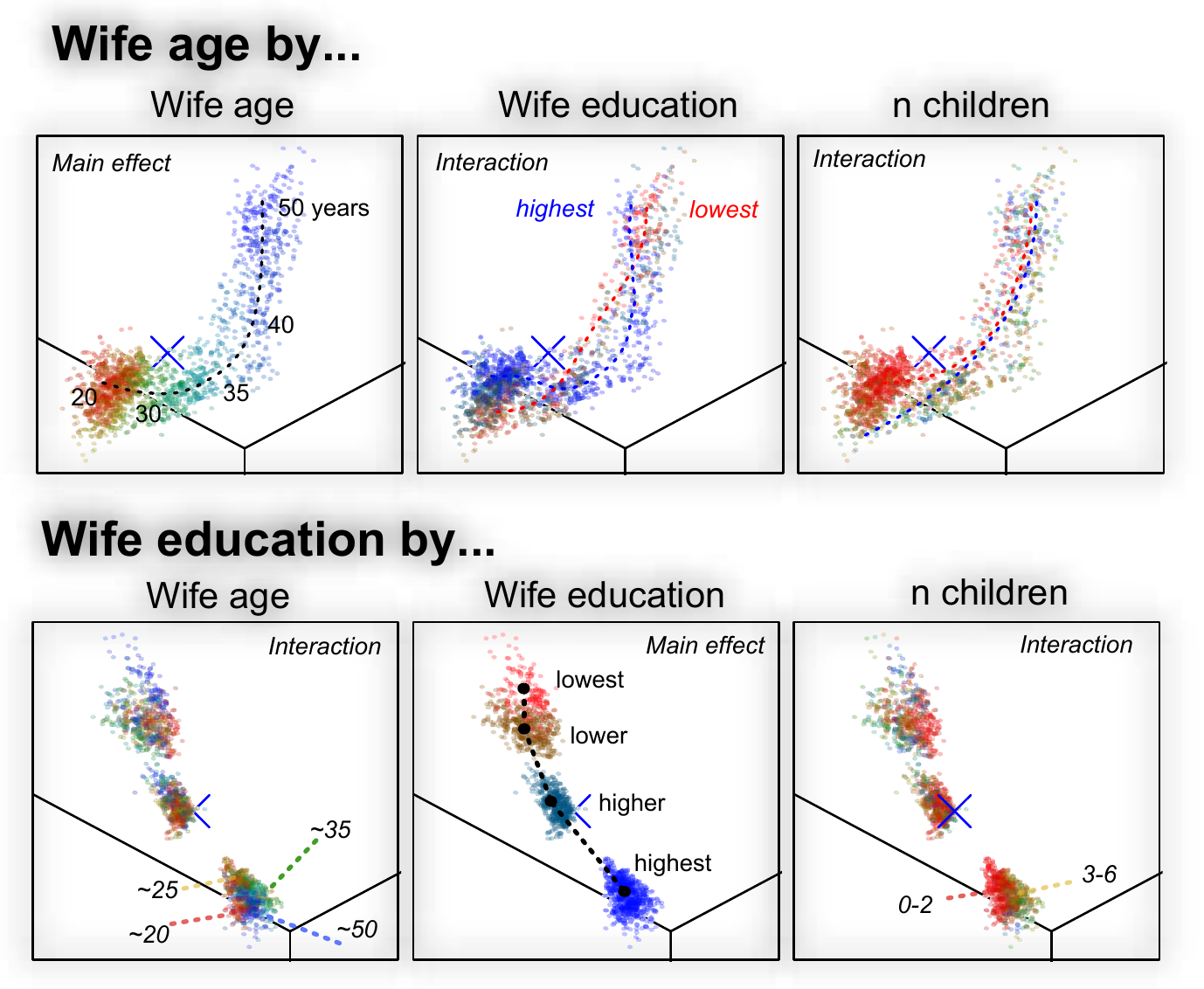}
\caption{Feature contributions for the three most important features plotted row-wise. Each plot is colored column-wise by corresponding feature values. Dash lines are drawn manually to assist interpretation of interactions.}
\label{cmc.2ndSimplex}
\end{figure*}

To search for higher order effects, similar to forest floor regression, simplex plots can in turn be colored by other features. In Figure \ref{cmc.2ndSimplex} the simplex plots of \emph{wives' age} and \emph{wives' education} was printed 3 times each. From left to right, color gradients illustrated respectively \emph{wives' age}, \emph{wives' education}, and lastly \emph{n\textunderscore children}. The simplexes in the diagonal reproduced the main effect coloring from Figure \ref{cmc.mainSimplex}, whereas other depicted simplexes possibly would detail 2\textsuperscript{nd} order interactions. E.g. \emph{wives' education} of Figure \ref{cmc.2ndSimplex} showed the four clusters, one for each education level. The distance from any point to its local cluster as a mix of higher order effects and a small noise component. It was found that wives with highest education aged 20 were predicted more likely to use contraception than when aged 25. Wives' with highest education and few children (red) preferred short term contraception over long term. As the features \emph{n\textunderscore children} and \emph{wives' age} are correlated, these will both interact with \emph{wives' education}, not only one.

\section{Discussion}
Forest floor is a methodology to visualize the mapping structure of a RF model using feature contributions. RF can be termed a predictive algorithmic model, designed to have a high predictive accuracy on the expense of model transparency \cite{Shmueli2010, Breimann2001}. RF could also be termed as data driven, as the model can adapt itself to the data with little guidance. The opposite is a theory driven model where the user manually choose an explicitly and clearly stated model to capture the data structure. A practical advantage of using RF, is when the user have little prior knowledge or theory on the subject. The majority of nonlinear machine learning algorithms models have in common, that the resulting model stated as an equation is fairly complex in the eyes of a human user. The complexity may be difficult to avoid if the model should be able to capture an unknown structure. But exactly when little prior theory is given, that is when the model should inspire the interpretation of the data structure. A dualistic approach is to choose both a perhaps linear explanatory model to interpret the system and a machine learning algorithm to get the most accurate predictions \cite{Shmueli2010}. Such an approach may leave a gap between users comprehension and the actual structure of the nonlinear model. If the user is far from understanding a certain data-structure any optimization cannot hardly evolve from brute trial-and-error searches such as grid search or ant-colony-optimization methods.

For nonlinear high-dimensional multivariate models, it is not straight forward to visualize the trained mapping function. The provided visualizations can be understood as slices or projections of the mapping structure. It appears that a given series of 2D and/or 3D projections can jointly explain the structure of a RF mapping surfacesa. The quantifiable goodness-of-visualization measure describes how well the variance of the full structure can be explained in the context of the provided feature axis. If a large component of feature contribution variance remains unexplained, there is likely an unaccounted interaction pattern associated with this feature. Thus an advantage of forest floor is, that it aids the user to learn what local interaction effects are not yet visualized. With feature contributions it is possible to make an interpretation of what variance is attributed main effects, second order effects or higher order effects. Feature contributions can be computed from the training set itself and thus do not extrapolate the training set. The training set is used to set boundaries for model structure, such that extrapolated and unrelated model structures are not visualized. Feature contributions can be combined with the out-of-bag concept allowing cross validation to avoid presenting an overfitted mapping structure. Visualizations of cross validated feature contributions appear less noisy.

Color gradients allowed to include one or two extra dimensions in an illustration thus otherwise limited of three dimension. Color gradients traversing entire mapping space was used to highlight selected latent dimensions in a series of main effect plots to pinpoint missing interactions. We perceive colors as a combination of three channels red, green and blue. Thus, it may seem possible to visualize three additional dimensions in colors. Nonetheless, the ranges of color saturation and brightness should be constrained to avoid indistinguishable grey color tones and to ensure a minimal contrast to the background. Such considerations, limited color gradients to provide only two additional dimensions at maximum. It was possible to summarize a high-dimensional structure with e.g. principal component analysis and apply color gradients along the first 2 loading vectors, such as in Figure \ref{cmc.mainSimplex}. In practice, we found a sequence of 1-dimensional color gradients best suited to uncover latent interaction structures in a RF model fit.

Feature contributions were first described in the context of RF regression, where a given feature can contribute either positively or negatively to a given prediction \cite{Kuz'min2011}. Next, the concept of feature contributions has previously been extended to classification, where the categorical majority vote labeling were replaced with numeric probability predictions \cite{Palczewska2014}. We have argued that these probabilistic predictions are confined in a prediction space defined the ($K-1$)-simplex, for model with $K$ classes. Any node in any tree will itself be a prediction and have a position in this space. We argue local increments are in fact vectors connecting nodes in the ($K-1$)-simplex space. The first local increment (the bootstrap increment) of any tree will be the vector connecting the class distribution of the training set to the class distribution of the root node. As the bootstrap increments will point randomly in any direction, the sum of a large number of such will approach the zero vector if no stratification is chosen. For stratification by true class, the bootstrap increments will connect the training set class distribution point in the ($K-1$)-simplex to the point in the ($K-1$)-simplex chosen by stratification. 

 The Gini loss function can be understood as maximizing the squared distance of node positions to the center of ($K-1$)-simplex (equal class probability). Therefore any split by Gini will place the daughter nodes the furthest from the center, weighted by node size. As the classification trees are fully grown, the terminal nodes of one pure class can only be positioned on the vertices of the simplex. In Figure \ref{cmc.mainSimplex} was shown that the distribution of classes in the training set will function effectively as the prior of the RF model. If the user do not expect to find the same class distribution in future predictions as in training set, this prior can be moved in the simplex by stratification during the bootstrap process. In Figure \ref{cmc.mainSimplex} the center blue cross marked that the average root node center was skewed towards class 1 (no contraception) as 42\% of the wives did not use any contraception. As class separation by the RF model was not strong the majority of predictions fall close to this prior base rate. In supplementary materials a RF model was trained with bootstrap stratification by true class such that the average root node is positioned in the center of the ($K-1$)-simplex and following predicted class probabilities were also centred around this point. Figure \ref{LIclass} depicted how any node-split will produce two new nodes with local increments in perfectly opposite direction. Thus, training set predictions will always be centred around this point.

Direct plotting of $K$ class probabilities requires $K-1$ dimensions. This is possible for 3 or 4 classes with 2D plot or 3D plot respectively. The context of feature values can only be included as one extra axis or as color gradients. We have shown that the axis of the ($K-1$)-simplex can be aligned such that only one axis is needed to visualize the feature contributions as seen in Figure \ref{cmc.mainAlign}. This frees 1 or 2 axis to provide an adequate feature value context. In such visualization each observation will be plotted one time for each predicted class probability. Colors can be used to distinguish the classes.

In a previous article we trained a molecular descriptor model with RF to predict protein permeation enhancement in an epithelial cell model (Caco-2) \cite{Welling2015}. A diagnostic tool was missed to address why such a model would be credible and to communicate intuitively the found pattern to fellow chemists/biologist with little knowledge of machine learning. We first stumbled upon feature contributions in the two articles \cite{Palczewska2014,Kuz'min2011} and experimented to plot these feature contributions against the feature values. The R package rfFC \cite{rfFC2013} provided the first computations of feature contributions and was an inspiration to the design of the forestFloor package \cite{forestFloor2015}. Hereafter we discovered partial dependence plots and sensitivity analysis \cite{Cortez2013,friedman2001}. Now in hindsight we can report the set of advantages to forest floor, especially the tracking of unaccounted interactions such that no strong interaction will be overlooked when visualizing the mapping structure.

The following citation by Friedman \cite{friedman2001} originates from an article from 2001 discussing the usefulness of partial dependence plots on nonlinear functions: \textit{"Given the general complexity of these generated targets as a function of their arguments, it is unlikely that one would ever be able to uncover their complete detailed functional form through a series of such partial dependence plots. The goal is to obtain an understandable description of some of the important aspects of the functional relationship."} \cite{friedman2001} \par
Indeed the structure of RF models can be highly complex and visualizations are unlikely to present every detail at once. Therefore a visualization tool-set should assist the user to navigate the mapping structure. This has been done by isolating the part of the model structure related to the data structure, by evaluating the goodness-of-visualization of a given plot, and by pointing to where locally in the model structure a sizable latent interaction is not yet visualized. Our goal is to present complex models as adequately detailed visualizations. In a RF model there will likely always be a baseline of random ripples in the mapping structure, that we do not expect to be able to reproduce. These ripples are partly filtered of by using the out-of-bag cross validated feature contributions. Other ripples occur due to biases of the RF algorithm. Especially does the RF model structure surface contain wave like curvature parallel to the feature axes due to the univariate step functions of RF, see RF surfaces in Supplementary Materials.

We predict that 4D projections of a third order interaction rarely would be needed for the RF algorithm. In supplementary materials we have provided a simulation suggesting that RF only poorly can fit interactions higher than second order even when trained on 10.000 observations without any noise. This can be explained as the RF algorithm is limited in its potential complexity as the algorithm only can perform univariate splits decided by an immediate loss function. Another algorithm such as rotation forest \cite{Rodriguez2006} is not limited to perform univariate splits and therefore better on such simulated tasks with higher order interactions. What initially was an interaction effect can be rearranged into a main effect by new combined features. Multivariate split methods are not compatible with forest floor, but they are compatible with the generic methods partial dependence plots and sensitivity analysis \cite{friedman2001,Cortez2013}.

\section{Conclusion}
Forest floor has extended the tool-box to visualize the mapping structure of RF models. The geometrical relationship between random forest models and feature contributions has been described. For RF multi-classification it was useful to understand the prediction space as a ($K-1$)-simplex probability space. Hereby the feature contributions can be interpreted as changes of predicted probability due to a given feature. A ($K-1$)-simplex prediction space can also visualize how the training set stratification affect RF predictions. Target class stratification is effective to modify the prior for the RF model.

We have emphasized that parts of a mapping structure which extrapolates the training set are irrelevant. To extract only the relevant mapping structure, feature contributions are computed only from the training set itself. Two new variants of feature contributions have been introduced to avoid inherent overfitting when using training set predictions. These variants of feature contributions are out-of-bag cross validated feature contributions, and n-fold cross validated feature contributions.

Feature contributions from a single feature can contain variance from main effects and/or interaction effects. A measure of goodness-of-visualization has been introduced to evaluate if the feature contributions of a given feature alone can be explained in the context of itself. If not, color gradients traversing the mapping space can be used to pin-point overlooked interactions within feature contributions and features. Sizable interactions can be visualized in two-way interaction plots in the context of two features and perhaps even a third feature as color gradient. Again a goodness-of-visualization can be computed and evaluated for such a visualization.

Ultimately, it is difficult to communicate a context of more than 2 or 3 dimensions + target dimension(s). Thus fourth order interactions would be difficult to visualize and communicate. Anyhow, such visualizations are likely not missed, as the random forest algorithm could not fit fourth order interactions well and had a poor efficiency already with third order interactions.

As forest floor can break down a RF model fit into effects attributed to each feature and assist to find adequate context to understand these effects. It is intended that RF no longer should be seen as a non interpretable model. Learned associations between features and targets should inspire new ideas of the underlying possible causality structure.

\clearpage
\end{multicols}

\clearpage
\bibliography{forstFloor}{}

\begin{thebibliography}{10}

\bibitem{Alhamdoosh2014}
Monther Alhamdoosh and Dianhui Wang.
\newblock Fast decorrelated neural network ensembles with random weights.
\newblock {\em Information Sciences}, 264(0):104 -- 117, 2014.
\newblock Serious Games.

\bibitem{rfFC2013}
Richard Marchese~Robinson Anna~Palczewska.
\newblock {\em rfFC: Random Forest Feature Contributions}, 2015.
\newblock R package version 1.0/r6.

\bibitem{Breimann2001}
Leo Breiman.
\newblock Statistical modeling: The two cultures.
\newblock {\em Statistical Science}, 16(3):pp. 199--215, 2001.

\bibitem{Cortez2009}
Paulo Cortez.
\newblock {UCI} machine learning repository, 2009.

\bibitem{Cortez2013}
Paulo Cortez and Mark~J. Embrechts.
\newblock Using sensitivity analysis and visualization techniques to open black
  box data mining models.
\newblock {\em Information Sciences}, 225(0):1 -- 17, 2013.

\bibitem{friedman2001}
Jerome~H Friedman.
\newblock Greedy function approximation: a gradient boosting machine.
\newblock {\em Annals of statistics}, pages 1189--1232, 2001.

\bibitem{icebox}
Alex Goldstein, Adam Kapelner, Justin Bleich, and Emil Pitkin.
\newblock Peeking inside the black box: Visualizing statistical learning with
  plots of individual conditional expectation.
\newblock {\em Journal of Computational and Graphical Statistics},
  24(1):44--65, 2015.

\bibitem{Hothorn2006}
Torsten Hothorn, Kurt Hornik, and Achim Zeileis.
\newblock Unbiased recursive partitioning: A conditional inference framework.
\newblock {\em Journal of Computational and Graphical statistics},
  15(3):651--674, 2006.

\bibitem{Kuz'min2011}
Victor~E. Kuz'min, Pavel~G. Polishchuk, Anatoly~G. Artemenko, and Sergey~A.
  Andronati.
\newblock Interpretation of qsar models based on random forest methods.
\newblock {\em Molecular Informatics}, 30(6-7):593--603, 2011.

\bibitem{Liaw2002}
Andy Liaw and Matthew Wiener.
\newblock Classification and regression by randomforest.
\newblock {\em R News}, 2(3):18--22, 2002.

\bibitem{Lim1987}
Tjen-Sien Lim.
\newblock {UCI} machine learning repository, 1987.

\bibitem{Lim2000}
Tjen-Sien Lim, Wei-Yin Loh, and Yu-Shan Shih.
\newblock A comparison of prediction accuracy, complexity, and training time of
  thirty-three old and new classification algorithms.
\newblock {\em Machine Learning}, 40(3):203--228, 2000.

\bibitem{Liu2014}
Sheng Liu, Shamitha Dissanayake, Sanjay Patel, Xin Dang, Todd Mlsna, Yixin
  Chen, and Dawn Wilkins.
\newblock Learning accurate and interpretable models based on regularized
  random forests regression.
\newblock {\em BMC Systems Biology}, 8(Suppl 3):S5, 2014.

\bibitem{Maree2005}
Raphael Maree, Pierre Geurts, Justus Piater, and Louis Wehenkel.
\newblock Random subwindows for robust image classification.
\newblock In {\em Computer Vision and Pattern Recognition, 2005. CVPR 2005.
  IEEE Computer Society Conference on}, volume~1, pages 34--40. IEEE, 2005.

\bibitem{OBrien2008}
Deirdre~B. O'Brien, Maya~R. Gupta, and Robert~M. Gray.
\newblock Cost-sensitive multi-class classification from probability estimates.
\newblock In {\em Proceedings of the 25th International Conference on Machine
  Learning}, ICML '08, pages 712--719, New York, NY, USA, 2008. ACM.

\bibitem{Palczewska2014}
Anna Palczewska, Jan Palczewski, Richard Marchese~Robinson, and Daniel Neagu.
\newblock Interpreting random forest classification models using a feature
  contribution method.
\newblock In Thouraya Bouabana-Tebibel and Stuart~H. Rubin, editors, {\em
  Integration of Reusable Systems}, volume 263 of {\em Advances in Intelligent
  Systems and Computing}, pages 193--218. Springer International Publishing,
  2014.

\bibitem{sklearn}
F.~Pedregosa, G.~Varoquaux, A.~Gramfort, V.~Michel, B.~Thirion, O.~Grisel,
  M.~Blondel, P.~Prettenhofer, R.~Weiss, V.~Dubourg, J.~Vanderplas, A.~Passos,
  D.~Cournapeau, M.~Brucher, M.~Perrot, and E.~Duchesnay.
\newblock Scikit-learn: Machine learning in {P}ython.
\newblock {\em Journal of Machine Learning Research}, 12:2825--2830, 2011.

\bibitem{R2015}
{R Core Team}.
\newblock {\em R: A Language and Environment for Statistical Computing}.
\newblock R Foundation for Statistical Computing, Vienna, Austria, 2015.

\bibitem{Rodriguez2006}
Juan~Jos{\'e} Rodriguez, Ludmila~I Kuncheva, and Carlos~J Alonso.
\newblock Rotation forest: A new classifier ensemble method.
\newblock {\em Pattern Analysis and Machine Intelligence, IEEE Transactions
  on}, 28(10):1619--1630, 2006.

\bibitem{Rstudio2015}
{RStudio Team}.
\newblock {\em RStudio: Integrated Development Environment for R}.
\newblock RStudio, Inc., Boston, MA, 2015.

\bibitem{Seligman2015}
Mark Seligman.
\newblock {\em Rborist: Extensible, Parallelizable Implementation of the Random
  Forest Algorithm}, 2015.
\newblock R package version 0.1-0.

\bibitem{Shmueli2010}
Galit Shmueli.
\newblock To explain or to predict?
\newblock {\em Statistical science}, pages 289--310, 2010.

\bibitem{Svetnik2003}
† Vladimir~Svetnik, *, † Andy~Liaw, † Christopher~Tong, ‡
  J.~Christopher~Culberson, §~Robert P.~Sheridan, , and Bradley~P. Feuston‡.
\newblock Random forest: A classification and regression tool for compound
  classification and qsar modeling.
\newblock {\em Journal of Chemical Information and Computer Sciences},
  43(6):1947--1958, 2003.
\newblock PMID: 14632445.

\bibitem{Welling2015}
Soeren~H. Welling, Line~K.H. Clemmensen, Stephen~T. Buckley, Lars Hovgaard,
  Per~B. Brockhoff, and Hanne~H.F. Refsgaard.
\newblock In silico modelling of permeation enhancement potency in caco-2
  monolayers based on molecular descriptors and random forest.
\newblock {\em European Journal of Pharmaceutics and Biopharmaceutics},
  94(0):152 -- 159, 2015.

\bibitem{forestFloor2015}
Soeren~Havelund Welling.
\newblock {\em forestFloor: Visualizes Random Forests with Feature
  Contributions}, 2015.
\newblock R package version 1.8.6.

\bibitem{Ranger2015}
M.~N. {Wright} and A.~{Ziegler}.
\newblock {ranger: A Fast Implementation of Random Forests for High Dimensional
  Data in C++ and R}.
\newblock {\em ArXiv e-prints}, August 2015.

\end{thebibliography}
\bibliographystyle{plain}
\end{document}